\documentclass[journal]{IEEEtran}
\ifCLASSINFOpdf
\else
\fi
\usepackage{url}

\usepackage[utf8]{inputenc}
\usepackage{xcolor}
\usepackage[colorlinks,pagebackref=true,breaklinks=true,citecolor=blue]{hyperref}
\usepackage{subcaption}
\usepackage[acronym,nohypertypes={acronym}]{glossaries}
\usepackage{color}
\usepackage{siunitx}
\usepackage{pgffor}
\DeclareMathOperator{\sign}{sign}
\usepackage[export]{adjustbox}
\usepackage{multirow}
\usepackage{bookmark}
\usepackage{amssymb}
\usepackage{amsthm}
\usepackage[numbers]{natbib}

\newcommand{\tabletopline}{\hline\noalign{\smallskip}}
\newcommand{\tablebottomline}{\noalign{\smallskip}\hline}
\newcommand{\tablemiddleline}{\noalign{\smallskip}\hline\noalign{\smallskip}}
\newcommand{\roundpvalue}[1]{\num[round-mode=places,round-minimum=.0001,round-precision=4,detect-all]{#1}}
\newcommand{\emphasized}[1]{\emph{\textcolor{red}{#1}}}
\newcommand{\emphbold}[1]{\textbf{#1}}
\newcommand{\emphstar}{\textbf{\color{blue}*}}
\newcommand{\pvalue}[2]{%
\ifthenelse{\equal{#2}{0}}%
{\emphasized}%
{}
{\ifthenelse{\lengthtest{#1pt > 0.05pt}}%
    {\roundpvalue{#1}}%
    {\emphbold{\roundpvalue{#1}}}%
    \ifthenelse{\lengthtest{#1pt > 0.01pt}}%
    {}{\emphstar}}}

\def\striplastbar#1{\striplastbara{#1}#1\end /\end\eend}
\def\striplastbara#1#2/\end#3\eend{\ifx\end#3\end#1\else#2\fi}
\newcommand{\changedir}[1]{\striplastbar{#1}__}

\newcommand\includeboundaries[2][]{\includegraphics[frame,#1]{#2}}

\newacronym[first={\glsentrytext{method}}]{method}{C}{??}
\newacronym[first={\glsentrytext{methodC}}]{methodC}{C$_{\mbox{\tiny C}}$}{??}
\newacronym[first={\glsentrytext{methodO}}]{methodO}{C$_{\mbox{\tiny O}}$}{??}
\newacronym{wsvm}{WSVM}{Weibull-calibrated Support Vector Machines}
\newacronym[first={\glsentrytext{wsvmO}}]{wsvmO}{WSVM$_{\mbox{\tiny O}}$}{??}
\newacronym[first={\glsentrytext{wsvmC}}]{wsvmC}{WSVM$_{\mbox{\tiny C}}$}{??}
\newacronym{pisvm}{PISVM}{Support Vector Machines with Probability of Inclusion}
\newacronym[first={\glsentrytext{pisvmO}}]{pisvmO}{PISVM$_{\mbox{\tiny O}}$}{??}
\newacronym[first={\glsentrytext{pisvmC}}]{pisvmC}{PISVM$_{\mbox{\tiny C}}$}{??}
\newacronym{evm}{EVM}{Extreme Value Machine}
\newacronym[first={\glsentrytext{evmO}}]{evmO}{EVM$_{\mbox{\tiny O}}$}{??}
\newacronym[first={\glsentrytext{evmC}}]{evmC}{EVM$_{\mbox{\tiny C}}$}{??}
\newacronym{ossvm}{OSSVM}{Open-Set Support Vector Machines}
\newacronym[first={\glsentrytext{ossvmC}}]{ossvmC}{SSVM$_{\mbox{\tiny C}}$}{??}
\newacronym[first={\glsentrytext{ossvmO}}]{ossvmO}{SSVM$_{\mbox{\tiny O}}$}{??}
\newacronym[first={\glsentrydesc{ocsvm} \cite[\glsentrytext{ocsvm};][]{Scholkopf2001a}}]{ocsvm}{OCSVM}{One-Class \glstext{svm}}
\newacronym{mcocsvm}{SVM$^{\mbox{\scriptsize OC}}$}{Multiclass One-Class \glsdesc{svm}}
\newacronym[first={\glsentrytext{mcocsvmO}}]{mcocsvmO}{SVM$^{\mbox{\scriptsize OC}}_{\mbox{\tiny O}}$}{??}
\newacronym[first={\glsentrytext{mcocsvmC}}]{mcocsvmC}{SVM$^{\mbox{\scriptsize OC}}_{\mbox{\tiny C}}$}{??}
\newacronym[first={\glsentrytext{ocbb}}]{ocbb}{OCBB}{one-class binary-based}
\newacronym{mcocbbsvm}{SVM$^{\mbox{\scriptsize \gls{ocbb}}}$}{multiclass binary-based \gls{ocsvm}}
\newacronym[first={\glsentrytext{mcocbbsvmC}}]{mcocbbsvmC}{SVM$^{\mbox{\scriptsize \gls{ocbb}}}_{\mbox{\tiny C}}$}{??}
\newacronym[first={\glsentrytext{mcocbbsvmO}}]{mcocbbsvmO}{SVM$^{\mbox{\scriptsize \gls{ocbb}}}_{\mbox{\tiny O}}$}{??}
\newacronym[first={\glsentrytext{mcocbbsvmOVO}}]{mcocbbsvmOVO}{SVM$^{\mbox{\scriptsize \gls{ocbb}}}_{\mbox{\tiny OVO}}$}{??}
\newacronym{mcsvdd}{SVDD$^{\mbox{\scriptsize OC}}$}{multiclass \glstext{svdd}}
\newacronym{mcsvddbb}{SVDD$^{\mbox{\scriptsize \gls{ocbb}}}$}{multiclass binary-based \glstext{svdd}}
\newacronym[first={\glsentrytext{mcsvddbbC}}]{mcsvddbbC}{SVDD$^{\mbox{\scriptsize \gls{ocbb}}}_{\mbox{\tiny C}}$}{??}
\newacronym[first={\glsentrytext{mcsvddbbO}}]{mcsvddbbO}{SVDD$^{\mbox{\scriptsize \gls{ocbb}}}_{\mbox{\tiny O}}$}{??}
\newacronym[first={\glsentrytext{mcsvddbbOVO}}]{mcsvddbbOVO}{SVDD$^{\mbox{\scriptsize \gls{ocbb}}}_{\mbox{\tiny OVO}}$}{??}
\newacronym[first={\glsentrydesc{svm} \cite[\glsentrytext{svm};][]{Cortes1995}}]{svm}{SVM}{Support Vector Machines}
\newacronym[first={\glsentrydesc{dnn} \cite[\glsentrytext{dnn};][]{LeCun2015}},firstplural={\glsentrydesc{dnn}s \cite[\glsentrytext{dnn}s;][]{LeCun2015}}]{dnn}{DNN}{deep neural network}
\newacronym[first={\glsentrytext{mcsvm}}]{mcsvm}{\gls{svm}}{??}
\newacronym[first={\glsentrytext{mcsvmC}}]{mcsvmC}{SVM$_{\mbox{\tiny C}}$}{??}
\newacronym[first={\glsentrytext{mcsvmO}}]{mcsvmO}{SVM$_{\mbox{\tiny O}}$}{??}
\newacronym[first={\glsentrydesc{knn} \cite[\glsentrytext{knn};][]{Bishop2006}}]{knn}{$k$NN}{$k$-Nearest Neighbors}
\newacronym{svm1vs}{1VS}{1-vs-Set Machine}
\newacronym[first={\glsentrytext{svm1vsO}}]{svm1vsO}{1VS$_{\mbox{\tiny O}}$}{??}
\newacronym[first={\glsentrytext{svm1vsC}}]{svm1vsC}{1VS$_{\mbox{\tiny C}}$}{??}
\newacronym{svmdbc}{DBC}{Decision Boundary Carving}
\newacronym[first={\glsentrytext{mcsvmdbc}}]{mcsvmdbc}{DBC}{\gls{mcfb} \glstext{svmdbc}}
\newacronym[first={\glsentrytext{mcsvmdbcO}}]{mcsvmdbcO}{DBC$_{\mbox{\tiny O}}$}{??}
\newacronym[first={\glsentrytext{mcsvmdbcC}}]{mcsvmdbcC}{DBC$_{\mbox{\tiny C}}$}{??}
\newacronym[first={\glsentrydesc{svdd} \cite[\glsentrytext{svdd};][]{Tax2004,Chang2013}}]{svdd}{SVDD}{Support Vector Data Description}
\newacronym[first={\glsentrytext{svddC}}]{svddC}{SVDD$_{\mbox{\tiny C}}$}{??}
\newacronym[first={\glsentrytext{svddO}}]{svddO}{SVDD$_{\mbox{\tiny O}}$}{??}
\newacronym[first={\glsentrytext{googlenet} \cite{Szegedy2015}}]{googlenet}{GoogLeNet}{??}
\newacronym{mcevm}{EVM}{Extreme Value Machine}

\newacronym{statsvm}{SVM}{Support Vector Machines}
\newacronym{statocsvm}{SVM$^{\mbox{\scriptsize OC}}$}{One-Class Support Vector Machines}
\newacronym{statsvdd}{SVDD$^{\mbox{\scriptsize OC}}$}{Support Vector Data Description}
\newacronym{statsvmdbc}{DBC}{Decision Boundary Carving}
\newacronym{statonevset}{1VS}{1-vs-Set Machine}
\newacronym{statwsvm}{WSVM}{Weibull-calibrated Support Vector Machines}
\newacronym{statpisvm}{PISVM}{Support Vector Machines with Probability of Inclusion}
\newacronym{statevm}{EVM}{Extreme Value Machine}
\newacronym{statssvm}{SSVM}{Specialized Support Vector Machines}

\newacronym[first={\glsentrytext{boat}}]{boat}{Boat}{\cite{Kuncheva2004}}
\newacronym[first={\glsentrytext{cone-torus}}]{cone-torus}{Cone-torus}{\cite{Kuncheva2004}}
\newacronym[first={\glsentrytext{four-gauss}}]{four-gauss}{Four-gauss}{\cite{Kuncheva2004}}
\newacronym[first={\glsentrytext{regular}}]{regular}{Regular}{\cite{Kuncheva2004}}

\newacronym[first={\glsentrytext{15-scenes} \glsentrydesc{15-scenes}}]{15-scenes}{15-Scenes}{\cite{Lazebnik2006}}
\newacronym[first={\glsentrytext{letter} \glsentrydesc{letter}}]{letter}{Letter}{\cite{Frey1991,Michie1994}}
\newacronym[first={\glsentrytext{auslan} \glsentrydesc{auslan}}]{auslan}{Auslan}{\cite{Kadous2002}}
\newacronym[first={\glsentrytext{caltech-256} \glsentrydesc{caltech-256}}]{caltech-256}{Caltech-256}{\cite{Griffin2007}}
\newacronym[first={\glsentrytext{aloi} \glsentrydesc{aloi}}]{aloi}{ALOI}{\cite{Geusebroek2005}}
\newacronym[first={\glsentrytext{imagenet}}]{imagenet}{ImageNet}{??}
\newacronym[first={\glsentrytext{imagenet2010}}]{imagenet2010}{ImageNet 2010}{\cite{Russakovsky2015}}
\newacronym[first={\glsentrytext{imagenet2012} \glsentrydesc{imagenet2012}}]{imagenet2012}{ImageNet 2012}{\cite{Russakovsky2015}}
\newacronym[first={\glsentrytext{cifar10}~\glsentrydesc{cifar10}}]{cifar10}{CIFAR10}{\cite{Krizhevsky2009}}
\newacronym[first={\glsentrytext{mnist}~\glsentrydesc{mnist}}]{mnist}{MNIST}{\cite{LeCun1998}}
\newacronym[first={\glsentrytext{krkopt}~\glsentrydesc{krkopt}}]{krkopt}{KRKOPT}{\cite{Olson2017,Bain1994}}
\newacronym[first={\glsentrytext{kddcup}~\glsentrydesc{kddcup}}]{kddcup}{KDDCUP}{\cite{Stolfo2000}}
\newacronym[first={\glsentrytext{movement}~\glsentrydesc{movement}}]{movement}{movement\_libras}{\cite{Olson2017}}
\newacronym[first={\glsentrytext{vowel}~\glsentrydesc{vowel}}]{vowel}{vowel}{\cite{Olson2017}}
\newacronym[first={\glsentrytext{led24}~\glsentrydesc{led24}}]{led24}{led24}{\cite{Olson2017}}
\newacronym[first={\glsentrytext{led7}~\glsentrydesc{led7}}]{led7}{led7}{\cite{Olson2017}}
\newacronym[first={\glsentrytext{mfeat-factors}~\glsentrydesc{mfeat-factors}}]{mfeat-factors}{mfeat-factors}{\cite{Olson2017}}
\newacronym[first={\glsentrytext{mfeat-fourier}~\glsentrydesc{mfeat-fourier}}]{mfeat-fourier}{mfeat-fourier}{\cite{Olson2017}}
\newacronym[first={\glsentrytext{mfeat-karhunen}~\glsentrydesc{mfeat-karhunen}}]{mfeat-karhunen}{mfeat-karhunen}{\cite{Olson2017}}
\newacronym[first={\glsentrytext{mfeat-morphological}~\glsentrydesc{mfeat-morphological}}]{mfeat-morphological}{mfeat-morphological}{\cite{Olson2017}}
\newacronym[first={\glsentrytext{mfeat-zernike}~\glsentrydesc{mfeat-zernike}}]{mfeat-zernike}{mfeat-zernike}{\cite{Olson2017}}
\newacronym[first={\glsentrytext{optdigits}~\glsentrydesc{optdigits}}]{optdigits}{optdigits}{\cite{Olson2017}}
\newacronym[first={\glsentrytext{pendigits}~\glsentrydesc{pendigits}}]{pendigits}{pendigits}{\cite{Olson2017}}
\newacronym[first={\glsentrytext{yeast}~\glsentrydesc{yeast}}]{yeast}{yeast}{\cite{Olson2017}}
\newacronym[first={\glsentrydesc{pmlb} \cite[\glsentrytext{pmlb};][]{Olson2017}}]{pmlb}{PMLB}{Penn Machine Learning Benchmark}

\newacronym{rbf}{RBF}{Radial Basis Function}
\newacronym{cap}{CAP}{Compact Abating Probability}
\newacronym{evt}{EVT}{Extreme Value Theory}
\newacronym{smo}{SMO}{Sequential Minimal Optimization}
\newacronym{wss}{WSS}{Working Set Selection}
\newacronym[first={\glsentrydesc{kkt} \cite[\glsentrytext{kkt};][]{Bishop2006}}]{kkt}{KKT}{Karush-Kuhn-Tucker}
\newacronym[first={\glsentrydesc{pca} \cite[\glsentrytext{pca};][]{Tipping1999}}]{pca}{PCA}{Principal Component Analysis}
\newacronym{tst}{TST}{Generalized T-Student}
\newacronym{rq}{RQ}{Rational Quadratic}
\newacronym{imq}{IMQ}{Inverse Multiquadric}
\newacronym[first={\glsentrydesc{grbf}}]{grbf}{RBF}{Generalized \glstext{rbf}}

\newacronym{cd}{CD}{Critical Difference}

\newacronym[first={\glsentrytext{ova}}]{ova}{one-vs-all}{??}
\newacronym[first={\glsentrytext{ovo}}]{ovo}{one-vs-one}{??}
\newacronym[first={\glsentrytext{dataset}},firstplural={\glsentrytext{dataset}s}]{dataset}{dataset}{??}
\newacronym[first={\glsentrytext{libsvm}}]{libsvm}{LibSVM}{??}
\newacronym[first={\glsentrytext{2d}}]{2d}{2-dimensional}{??}

\newacronym[first={\emph{\glsentrytext{ccv}}}]{ccv}{cross-class validation}{??}
\newacronym[first={\glsentrytext{openspace}}]{openspace}{open space}{??}
\newacronym[first={\glsentrytext{mcfb}}]{mcfb}{multiclass-from-binary}{multiclass-from-binary}
\newacronym[first={\glsentrytext{acs}},symbol={\mbox{ACS}}]{acs}{available classes}{available classes}
\newacronym[first={\emph{\glsentrytext{dr}}},firstplural={\emph{\glsentrytext{dr}s}}]{dr}{decision region}{??}
\newacronym{klos}{KLOS}{\emph{known labeled open space}}
\newacronym{plos}{PLOS}{\emph{positively labeled open space}}
\newacronym[first={\glsentrytext{gs}}]{gs}{grid search}{??}
\newacronym[first={\emph{\glsentrytext{ru}}}]{ru}{risk of the unknown}{??}
\newacronym[first={\emph{\glsentrytext{er}}}]{er}{empirical risk}{??}
\newacronym[first={\emph{\glsentrytext{tde}}},firstplural={\emph{\glsentrytext{tde}s}}]{tde}{training data error}{??}
\newacronym[first={\emph{\glsentrytext{osr}}},firstplural={\emph{\glsentrytext{osr}s}}]{osr}{open-space risk}{??}
\newacronym[first={\emph{\glsentrytext{ro}}},firstplural={\emph{\glsentrytext{ro}s}}]{ro}{reject option}{??}

\newacronym[symbol={\mbox{\glsentrytext{aks}}}]{aks}{AKS}{accuracy on known samples}
\glsunset{aks}
\newacronym[symbol={\mbox{\glsentrytext{fu}}},first={\glsentrytext{fu}}]{fu}{false unknown}{??}
\newacronym[symbol={\mbox{\glsentrytext{mis}}},first={\glsentrytext{mis}}]{mis}{misclassification}{??}
\newacronym[symbol={\mbox{\glsentrytext{aus}}}]{aus}{AUS}{accuracy on unknown samples}
\glsunset{aus}
\newacronym[symbol={\mbox{\glsentrytext{na}}}]{na}{NA}{normalized accuracy}
\newacronym[symbol={\mbox{\glsentrytext{harmonicNA}}}]{harmonicNA}{HNA}{harmonic normalized accuracy}
\newacronym[symbol={\mbox{\glsentrytext{harmonicNA}}},first={\gls{harmonicNA}}]{hna}{\gls{harmonicNA}}{??}
\newacronym[symbol={\mbox{\glsentrytext{osfm}}}]{osfm}{OSFM}{open-set f-measure}
\newacronym[first={\glsentrytext{fmeasure}},symbol={\mbox{\glsentrytext{fmeasure}}}]{fmeasure}{\mbox{f-measure}}{??}
\newacronym[first={\glsentrytext{precision}},symbol={\mbox{\glsentrytext{precision}}}]{precision}{precision}{??}
\newacronym[first={\glsentrytext{precisionM}},symbol={\mbox{\glsentrytext{precisionM}}}]{precisionM}{precision$_M$}{??}
\newacronym[first={\glsentrytext{precisionm}},symbol={\mbox{\glsentrytext{precisionm}}}]{precisionm}{precision$_{\mu}$}{??}
\newacronym[first={\glsentrytext{recall}},symbol={\mbox{\glsentrytext{recall}}}]{recall}{recall}{??}
\newacronym[first={\glsentrytext{recallM}},symbol={\mbox{\glsentrytext{recallM}}}]{recallM}{recall$_M$}{??}
\newacronym[first={\glsentrytext{recallm}},symbol={\mbox{\glsentrytext{recallm}}}]{recallm}{recall$_{\mu}$}{??}
\newacronym[symbol={\mbox{\glsentrytext{osfmM}}}]{osfmM}{OSFM$_M$}{macro-averaging open-set f-measure}
\newacronym[first={\glsentrytext{mafm}}]{mafm}{\gls{osfmM}}{??}
\newacronym[symbol={\mbox{\glsentrytext{osfmm}}}]{osfmm}{OSFM$_{\mu}$}{micro-averaging open-set f-measure}
\newacronym[first={\glsentrytext{mifm}}]{mifm}{\gls{osfmm}}{??}
\newacronym[symbol={\mbox{\glsentrytext{fmM}}}]{fmM}{FM$_M$}{macro-averaging f-measure}
\newacronym[first={\glsentrytext{bbmafm}}]{bbmafm}{\gls{fmM}}{??}
\newacronym[symbol={\mbox{\glsentrytext{fmm}}}]{fmm}{FM$_{\mu}$}{micro-averaging f-measure}
\newacronym[first={\glsentrytext{bbmifm}}]{bbmifm}{\gls{fmm}}{??}
\newacronym[first={\glsentrytext{tp}},symbol={\mbox{TP}}]{tp}{true positive}{\mbox{TP$_i$}}
\newacronym[first={\glsentrytext{fp}},symbol={\mbox{FP}}]{fp}{false positive}{\mbox{FP$_i$}}
\newacronym[first={\glsentrytext{fn}},symbol={\mbox{FN}}]{fn}{false negative}{\mbox{FN$_i$}}
\newacronym[first={\glsentrytext{tn}},symbol={\mbox{TN}}]{tn}{true negative}{\mbox{TN$_i$}}
\newacronym{AKS}{AKS}{accuracy on known samples}
\newacronym{AUS}{AUS}{accuracy on unknown samples}
\newacronym{NA}{NA}{normalized accuracy}
\newacronym{HNA}{HNA}{harmonic normalized accuracy}
\newacronym{OSFMM}{OSFM$_M$}{macro-averaging open-set f-measure}
\newacronym{OSFMm}{OSFM$_{\mu}$}{micro-averaging open-set f-measure}
\newacronym{FMM}{FM$_M$}{macro-averaging f-measure}
\newacronym{FMm}{FM$_{\mu}$}{micro-averaging f-measure}

\newacronym[first={\glsentrytext{number-of-datasets}}]{number-of-datasets}{nineteen}{??}

\newcommand{\figPlottingkernel}{
  \begin{figure*}
    \centering
    \hspace{2cm}
    \begin{subfigure}[t]{0.30\linewidth}
      \centering
      \includegraphics[width=0.95\textwidth]{\changedir{figsR2/}figBoundaries__boat_forboundaries__color}
      \caption*{\glstext{boat} \gls{dataset} with 3 classes: red (the central class to the left), green (the central class to the right), and blue (the class with the ring shape).}
      \label{fig:figBoundaries__boat_forboundaries__color}
    \end{subfigure}\hfill%
    \foreach \size/\filename/\captiontext in {%
      /figPlottingkernel3d__boat_forboundaries__cls01__01__gammaexp+04/\caption{Class 1 (red). \mbox{$b = -0.832$}.},
      /figPlottingkernel3d__boat_forboundaries__cls02__01__gammaexp+04/\caption{Class 2 (green). \mbox{$b = -0.86$}.},
      /figPlottingkernel3d__boat_forboundaries__cls03__01__gammaexp+04/\caption{Class 3 (blue). \mbox{$b = +0.594$}.}%
    }{%
      \ifthenelse{\equal{\size}{}}{\renewcommand\size{0.50}}{}%
      \begin{subfigure}[t]{\size\linewidth}
        \centering
        \includegraphics[width=0.95\textwidth]{\changedir{figs/}\filename}
        \captiontext
        \label{fig:\filename}
      \end{subfigure}\hfill%
    }
    \caption{%
      Behavior of the \glsdesc{svm} (\glstext{svm}) with a \glsfirst{rbf} kernel on a \gls{2d} synthetic \gls{dataset}.
      Image on the top-left depicts the \gls{boat} \gls{dataset} \cite{Kuncheva2004}.
      Figures (\subref{fig:figPlottingkernel3d__boat_forboundaries__cls01__01__gammaexp+04})--(\subref{fig:figPlottingkernel3d__boat_forboundaries__cls03__01__gammaexp+04}) correspond to the red, green and blue classes of the \gls{boat} \gls{dataset}, respectively.
      $x$, $y$ axes in Figures (\subref{fig:figPlottingkernel3d__boat_forboundaries__cls01__01__gammaexp+04})--(\subref{fig:figPlottingkernel3d__boat_forboundaries__cls03__01__gammaexp+04}) represent the two features of the \gls{dataset} (the \gls{dataset} is normalized between 0 and 1).
      $z$ axis shows the value of the decision function $\sum_{i=1}^{m} y_i \alpha_i K(\mathbf{x}_i, \mathbf{x}) + b$ (Equation \ref{eq:decision-function-kernel} without $\sign$ function) and the colored lines in the walls depict the point 0, that separates the positive class from the negative one (equivalent to the $\sign$ function of Equation \ref{eq:decision-function-kernel}).
      Note in Figure (\subref{fig:figPlottingkernel3d__boat_forboundaries__cls03__01__gammaexp+04}) that an unbounded region of the feature space remains in the positive side, as $b > 0$ and $f(x, y) \approx b$ for ($x$, $y$) points far away from support vectors.
    }\label{fig:figPlottingkernel}
  \end{figure*}
}

\newcounter{figBoundariesCounter}
\newcounter{figBoundariesCounterTwo}
\newcommand{\figBoundaries}[2]{
  \setcounter{figBoundariesCounter}{-1}
  \setcounter{figBoundariesCounterTwo}{0}
  \begin{figure}
    \tiny
    \centering
    \foreach \filename/\classifier in {%
      /\Gls{dataset},
      __mcsvm_ova_gsic_highGamma_fixedC/\gls{mcsvm},
      __mcocsvm_ova_gsic/\gls{mcocsvm},
      __mcsvdd_ova_gsic/\gls{mcsvdd},
      __mcsvmdbc_ova_gsic/\gls{mcsvmdbc},
      __svm1vsll_ovx_gsec/\gls{svm1vs},
      __wsvm_ovx_gsec/\gls{wsvm},
      __pisvm_ovx_gseo/\gls{pisvm},
      __mcevm_ovx_gseo/\gls{evm},
      __mcossvm_ova_gsic/\gls{ossvm}%
    }{%
      \begin{subfigure}[t]{0.24\linewidth}
        \centering
        \includeboundaries[width=\textwidth]{\changedir{figsR2/}figBoundaries__#1_forboundaries\filename__color}
        \ifthenelse{\equal{\filename}{}}{\caption*}{\caption}{\footnotesize\classifier}
        \label{subfig:#1\filename}
      \end{subfigure}
      \addtocounter{figBoundariesCounter}{1}%
      \ifthenelse{\value{figBoundariesCounter}=3}{\addtocounter{figBoundariesCounterTwo}{1}\setcounter{figBoundariesCounter}{0}\\}{}%
      \ifthenelse{\value{figBoundariesCounterTwo}=3}{\setcounter{figBoundariesCounterTwo}{0}\setcounter{figBoundariesCounter}{-1}}{}%
    }
    \caption{%
      \Glspl{dr} for the #2 \gls{dataset}.
      Non-white regions represent the region in which a test sample would be classified as belonging to the same class of the samples with the same color.
      All samples in the white regions would be classified as unknown.
    }\label{fig:#1}
  \end{figure}
}

\newcommand{\thousands}[1]{\num[group-separator={,},round-precision=0,round-mode=places]{#1}}

\newcommand{\tableBaselinesSummary}{
  \begin{table}
    \centering
    \resizebox{\linewidth}{!}{
      \begin{tabular}{rlccccc}
        \tabletopline
        \multicolumn{2}{c}{\multirow{2}{*}{\textbf{Method}}} & \textbf{\glstext{rbf}} & \textbf{\glstext{evt}} & \textbf{One-class} & \textbf{Bounded}        \\
                                      &                      & \textbf{kernel}        & \textbf{based}         & \textbf{based}     & \textbf{\Glstext{klos}} \\
        \tablemiddleline
        \textbf{\glstext{mcsvm}}      & \citet{Chang2011}    & \checkmark             &                        &                    &                         \\
        \textbf{\glstext{mcocsvm}}    & \citet{Pritsos2013}  & \checkmark             &                        & \checkmark         & \checkmark              \\
        \textbf{\glstext{mcsvdd}}     & \citet{Tax2004}      & \checkmark             &                        & \checkmark         & \checkmark              \\
        \textbf{\glstext{mcsvmdbc}}   & \citet{Costa2014}    & \checkmark             &                        &                    &                         \\
        \textbf{\glstext{svm1vs}}     & \citet{Scheirer2013} &                        &                        &                    &                         \\
        \textbf{\glstext{wsvm}}       & \citet{Scheirer2014} & \checkmark             & \checkmark             & \checkmark         & \checkmark              \\
        \textbf{\glstext{pisvm}}      & \citet{Jain2014}     & \checkmark             & \checkmark             &                    &                         \\
        \textbf{\glstext{evm}}        & \citet{Rudd2018}     &                        & \checkmark             &                    & \checkmark              \\
        \textbf{\glstext{ossvm}}      & (Proposed method)    & \checkmark             &                        &                    & \checkmark              \\
        \tablebottomline
      \end{tabular}
    }
    \caption{General characteristics of the methods compared in the experiments.}
    \label{tab:baselines-summary}%
  \end{table}
}
\newcommand{\tableDatasetsInfo}{
  \begin{table}
    \centering
    \resizebox{\linewidth}{!}{
      \begin{tabular}{crrrrrr}
        \tabletopline
        \multirow{2}{*}{\textbf{\Glstext{dataset}}} & \multirow{2}{*}{\textbf{\# classes}} & \multirow{2}{*}{\textbf{\# samples}} & \multirow{2}{*}{\textbf{\# features}} & \multicolumn{3}{c}{\textbf{\# samples/class}}                 \\
                                                    &                                      &                                      &                                       & \textbf{mean}           & \textbf{min}     & \textbf{max}     \\
        \tablemiddleline
        \textbf{\glstext{yeast}}                    & \thousands{9}                        & \thousands{1479}                     & \thousands{8}                         & \thousands{164.333333}  & \thousands{20}   & \thousands{463}  \\
        \textbf{\glstext{mfeat-morphological}}      & \thousands{10}                       & \thousands{2000}                     & \thousands{6}                         & \thousands{200.000000}  & \thousands{200}  & \thousands{200}  \\
        \textbf{\glstext{mfeat-zernike}}            & \thousands{10}                       & \thousands{2000}                     & \thousands{47}                        & \thousands{200.000000}  & \thousands{200}  & \thousands{200}  \\
        \textbf{\glstext{mfeat-karhunen}}           & \thousands{10}                       & \thousands{2000}                     & \thousands{64}                        & \thousands{200.000000}  & \thousands{200}  & \thousands{200}  \\
        \textbf{\glstext{mfeat-fourier}}            & \thousands{10}                       & \thousands{2000}                     & \thousands{76}                        & \thousands{200.000000}  & \thousands{200}  & \thousands{200}  \\
        \textbf{\glstext{mfeat-factors}}            & \thousands{10}                       & \thousands{2000}                     & \thousands{216}                       & \thousands{200.000000}  & \thousands{200}  & \thousands{200}  \\
        \textbf{\glstext{led7}}                     & \thousands{10}                       & \thousands{3200}                     & \thousands{7}                         & \thousands{320.000000}  & \thousands{270}  & \thousands{341}  \\
        \textbf{\glstext{led24}}                    & \thousands{10}                       & \thousands{3200}                     & \thousands{24}                        & \thousands{320.000000}  & \thousands{296}  & \thousands{337}  \\
        \textbf{\glstext{optdigits}}                & \thousands{10}                       & \thousands{5620}                     & \thousands{64}                        & \thousands{562.000000}  & \thousands{554}  & \thousands{572}  \\
        \textbf{\glstext{pendigits}}                & \thousands{10}                       & \thousands{10992}                    & \thousands{16}                        & \thousands{1099.200000} & \thousands{1055} & \thousands{1144} \\
        \textbf{\glstext{cifar10}}                  & \thousands{10}                       & \thousands{50000}                    & \thousands{192}                       & \thousands{5000.000000} & \thousands{5000} & \thousands{5000} \\
        \textbf{\glstext{mnist}}                    & \thousands{10}                       & \thousands{55000}                    & \thousands{1024}                      & \thousands{5500.000000} & \thousands{4987} & \thousands{6179} \\
        \textbf{\glstext{vowel}}                    & \thousands{11}                       & \thousands{990}                      & \thousands{13}                        & \thousands{90.000000}   & \thousands{90}   & \thousands{90}   \\
        \textbf{\glstext{movement}}                 & \thousands{15}                       & \thousands{360}                      & \thousands{90}                        & \thousands{24.000000}   & \thousands{24}   & \thousands{24}   \\
        \textbf{\glstext{15-scenes}}                & \thousands{15}                       & \thousands{4485}                     & \thousands{1000}                      & \thousands{299.000000}  & \thousands{210}  & \thousands{410}  \\
        \textbf{\glstext{krkopt}}                   & \thousands{18}                       & \thousands{28056}                    & \thousands{6}                         & \thousands{1558.666667} & \thousands{27}   & \thousands{4553} \\
        \textbf{\glstext{letter}}                   & \thousands{26}                       & \thousands{20000}                    & \thousands{16}                        & \thousands{769.230769}  & \thousands{734}  & \thousands{813}  \\
        \textbf{\glstext{kddcup}}                   & \thousands{32}                       & \thousands{10237}                    & \thousands{41}                        & \thousands{319.906250}  & \thousands{11}   & \thousands{500}  \\
        \textbf{\glstext{auslan}}                   & \thousands{95}                       & \thousands{146949}                   & \thousands{22}                        & \thousands{1546.831579} & \thousands{1390} & \thousands{1938} \\
        \textbf{\glstext{caltech-256}}              & \thousands{256}                      & \thousands{29780}                    & \thousands{1000}                      & \thousands{116.328125}  & \thousands{80}   & \thousands{800}  \\
        \textbf{\glstext{imagenet}}                 & \thousands{360}                      & \thousands{190780}                   & \thousands{1024}                      & \thousands{529.944444}  & \thousands{520}  & \thousands{530}  \\
        \textbf{\glstext{aloi}}                     & \thousands{1000}                     & \thousands{108000}                   & \thousands{128}                       & \thousands{108.000000}  & \thousands{108}  & \thousands{108}  \\
        \tablebottomline
      \end{tabular}
    }
    \caption{General characteristics of the \glspl{dataset} employed in the experiments.}
    \label{tab:datasets}%
  \end{table}
}

\newcommand{\tablePercentageNegativeBias}{
  \begin{table}
    \centering
    \begin{tabular}{cccccc}
      \tabletopline
                                             & \textbf{\glstext{svm} \glstext{ova}} & \textbf{\glstext{svm} \glstext{ovo}} \\
      \tablemiddleline
      \textbf{\glstext{yeast}}               & 93.33\%                              & 26.40\%                              \\
      \textbf{\glstext{mfeat-morphological}} & 98.33\%                              & 39.66\%                              \\
      \textbf{\glstext{mfeat-zernike}}       & 98.33\%                              & 47.11\%                              \\
      \textbf{\glstext{mfeat-karhunen}}      & 98.89\%                              & 62.13\%                              \\
      \textbf{\glstext{mfeat-fourier}}       & 97.78\%                              & 57.70\%                              \\
      \textbf{\glstext{mfeat-factors}}       & 99.44\%                              & 41.82\%                              \\
      \textbf{\glstext{led7}}                & 100.0\%                              & 57.36\%                              \\
      \textbf{\glstext{led24}}               & 100.0\%                              & 50.00\%                              \\
      \textbf{\glstext{optdigits}}           & 99.44\%                              & 69.25\%                              \\
      \textbf{\glstext{pendigits}}           & 96.11\%                              & 64.25\%                              \\
      \textbf{\glstext{cifar10}}             & 100.0\%                              & 38.70\%                              \\
      \textbf{\glstext{mnist}}               & 99.44\%                              & 43.52\%                              \\
      \textbf{\glstext{vowel}}               & 98.33\%                              & 67.10\%                              \\
      \textbf{\glstext{movement}}            & 100.0\%                              & 49.96\%                              \\
      \textbf{\glstext{15-scenes}}           & 99.00\%                              & 56.92\%                              \\
      \textbf{\glstext{krkopt}}              & 95.67\%                              & 40.25\%                              \\
      \textbf{\glstext{letter}}              & 99.67\%                              & 55.75\%                              \\
      \textbf{\glstext{kddcup}}              & 99.33\%                              & 49.79\%                              \\
      \textbf{\glstext{auslan}}              & 99.00\%                              & 42.50\%                              \\
      \textbf{\glstext{caltech-256}}         & 98.00\%                              & 48.75\%                              \\
      \textbf{\glstext{imagenet}}            & 99.33\%                              & 45.08\%                              \\
      \textbf{\glstext{aloi}}                & 97.67\%                              & 52.50\%                              \\
      \tablemiddleline
      \textbf{Mean}                          & 98.50\%                              & 50.30\%                              \\
      \tablebottomline
    \end{tabular}
    \caption{Percentage of binary classifiers with negative bias term per \gls{dataset} for \glstext{ova} and \glstext{ovo} approaches.}
    \label{tab:correct-bias-term-frequency}
  \end{table}
}

\newtheorem{theorem}{Theorem}
\newtheorem{proposition}{Proposition}

\begin{document}
%
\title{Open-Set Support Vector Machines}
%
%
%

\author{Pedro~Ribeiro~Mendes~J{\'u}nior,
  Terrance~E.~Boult,~\IEEEmembership{Fellow,~IEEE,}\\
  Jacques~Wainer,
  and~Anderson~Rocha,~\IEEEmembership{Fellow,~IEEE}
\thanks{P. R. Mendes~Júnior, J. Wainer, and A. Rocha are with the RECOD Lab., Institute of Computing (IC), University of Campinas (UNICAMP), Av.\@ Albert Einstein, 1251, Campinas, SP, 13083-852, Brazil e-mail: (see \url{https://pedrormjunior.github.io}).}
\thanks{T. E. Boult is with VAST Lab., Department of Computer Science, Engineering Building, University of Colorado Colorado Springs (UCCS), 1420 Austin Bluffs Parkway, Colorado Springs, CO 80918, USA.}
\thanks{Manuscript received November 13, 2019; revised April 17, 2021.}}

%
%

\markboth{IEEE Transactions on Systems, Man, and Cybernetics: Systems}%
{Mendes~Júnior \MakeLowercase{\textit{et al.}}: Open-Set Support Vector Machines}
%



\maketitle

\begin{abstract}
  Often, when dealing with real-world recognition problems, we do not need, and often cannot have, knowledge of the entire set of possible classes that might appear during operational testing.
  In such cases, we need to think of robust classification methods able to deal with the ``unknown'' and properly reject samples belonging to classes never seen during training.
  Notwithstanding, existing classifiers to date were mostly developed for the closed-set scenario, i.e., the classification setup in which it is assumed that all test samples belong to one of the classes with which the classifier was trained.
  In the open-set scenario, however, a test sample can belong to none of the known classes and the classifier must properly reject it by classifying it as unknown.
  In this work, we extend upon the well-known Support Vector Machines (SVM) classifier and introduce the Open-Set Support Vector Machines (OSSVM), which is suitable for recognition in open-set setups.
  OSSVM balances the empirical risk and the risk of the unknown and ensures that the region of the feature space in which a test sample would be classified as known (one of the known classes) is always bounded, ensuring a finite risk of the unknown.
  In this work, we also highlight the properties of the SVM classifier related to the open-set scenario, and provide necessary and sufficient conditions for an RBF SVM to have bounded open-space risk.
\end{abstract}


%
\IEEEpeerreviewmaketitle

\section{Introduction}\label{sec:introduction}
%
%
%
%

\IEEEPARstart{M}{achine} learning literature is rich with works proposing classifiers for closed-set pattern recognition, with well-known examples including \gls{knn}, Random Forests \cite{Breiman2001}, \gls{svm}, and \glspl{dnn}.
These classifiers were inherently designed to work in closed-set scenarios, i.e., scenarios in which all test samples must belong to a class used in training.
What happens when the test sample belongs to a class not seen at training time?
Consider a digital forensic scenario---e.g., source-camera attribution \cite{Costa2014}, printer identification \cite{Ferreira2017}---in which law officials want to verify that a particular artifact (e.g., a digital photo or a printed page) did originate from one of a few suspect devices.
The suspected devices are the classes of interest, and a classifier can be trained on many examples of artifacts from these devices and from other non-suspected devices.
When assigning the source for the particular artifact in question, the classifier must be aware that if the artifact is from an unknown region of the feature space---possibly far away from the training data---it cannot be assigned to one of the suspected devices, even if examples of those devices are the closest to the artifact.
The classifier must be allowed to declare that the example does not belong to any of the classes it was trained on.

One possible way to address recognition in an open-set scenario is to use a closed-set classifier, obtain a similarity score---or simply the distance in the feature space---to the most likely class, apply a threshold on that similarity score aiming at classifying as unknown any test sample whose similarity score is below a specified threshold \cite{Dubuisson1993,Muzzolini1998,Moeini2017}.
\citet{MendesJunior2017} showed that when applying thresholds to the ratio of distances instead of distance themselves results in better performance in open-set scenarios.
Furthermore, recent works have shown theoretical and experimental inconsistencies on employing thresholded softmax probability scores of neural networks for open-set rejection in face recognition problems \cite{Liu2017,Wen2019}.

Instead of using similarity-based algorithms, another alternative is to exploit kernel-based algorithms such as \glsfirst{svdd} and one-class classifiers \cite{Scholkopf2001a}---applied to the entire training set as a rejection function \cite{Cortes2016}.
This approach is sometimes called \textit{classification with abstention}.
The idea is to have an initial rejection phase that predicts if the input belongs to one of the training classes or not (known or unknown).
In the former case, a second phase is performed with any sort of multiclass classifier aiming at choosing the correct class.

Another alternative method relies on having a binary rejection function for each of the known classes such that a test sample is classified as unknown when decisions are negative for every function.
This is the case for any \gls{mcfb} classifier based on the \gls{ova} approach \cite{Rocha2014}.
Some recent works have explored this idea \cite{Pritsos2013,Costa2014,Scheirer2014,Jain2014} making efforts at minimizing the \gls{plos} for every classifier that composes the \gls{mcfb} one.
In binary classification, \gls{plos} refers to the \gls{openspace} that receives positive classification.
\Gls{openspace} is the region of the feature space outside the support of the training samples \cite{Scheirer2013}.
In the multiclass level, a similar concept applies: \gls{klos}, i.e., the region of the feature space outside the support of the training samples in which a test sample would be classified as one of the known classes \cite{MendesJunior2017}.
In this class of methods, the potential of binary methods can be highly explored to multiclass open-set scenarios.
Furthermore, methods in this class can be adapted for multiple class recognition in open-set scenarios, as accomplished by \citet{Heflin2012}.

In this work, we propose the \gls{ossvm}, that falls into the last class of methods and receives its name due to its ability to bound the \gls{plos} for every classification in the binary level, consequently, bounding \gls{klos} as well---when \gls{ova} is applied.
\gls{ossvm} relies on the optimization of the bias term with \gls{rbf} kernel taking advantage of the following property we demonstrate in this work: \gls{svm} with \gls{rbf} kernel bounds the \gls{plos} if and only if the bias term is negative.

Along with this work, we have evaluated multiple implementations of open-set methods for different \glspl{dataset}.
For all those methods we have employed the \emph{open-set grid search}, as it was showed by \citet{MendesJunior2019} that it works better over the traditional form of grid search in open-set scenarios.\footnote{The \emph{open-set grid search} is a generalization of both the \gls{ccv} of \citet{Jain2014} and the \emph{parameter optimization} of \citet{MendesJunior2017}.  \citet{MendesJunior2017} have suggested their parameter optimization as a general form of grid search, which were later on formalized and extensively evaluated \citep{MendesJunior2019}.}

The remaining of this work is organized as follows.
In Section \ref{sec:related-work}, we discuss some of the most important previous work in open-set recognition.
In Section \ref{sec:ssvm}, we introduce the \gls{ossvm} while, in Section \ref{sec:experiments}, we present the experiments that validate the proposed method.
Finally, in Section \ref{sec:conclusion}, we present the conclusions and future work.

\section{Related Work}\label{sec:related-work}

In this section, we review recent works that explicitly deal with open-set recognition in the literature, including some base works for them.
We note that other insights presented in many existing works can be somehow extended to be employed for the open-set scenario.
Most of those works, however, did not perform the experiments with appropriate open-set recognition setup.

\citet{Heflin2012} and \citet{Pritsos2013} presented a multiclass \gls{svm} classifier based on the \gls{ocsvm}.
For each of the training classes, they fit an \gls{ocsvm}.
In the prediction phase, all $n$ \glspl{ocsvm} classify the test sample, in which $n$ is the number of \gls{acs} for training.
The test sample is classified as the class in which its \gls{ocsvm} classified as positive.
When no \gls{ocsvm} classifies the test sample as positive, it is classified as unknown.
\citet{Heflin2012} extends the idea to multiple class classification, by allowing more than one \glspl{ocsvm} to classify the example as positive; in this case, the example receives as labels all classes whose corresponding \glspl{ocsvm} classify it as positive.
Differently, \citet{Pritsos2013} choose the more confident classifier among the ones that classify as positive.
In those works, \gls{ocsvm} is used with the \gls{rbf} kernel, which allows bounding the \gls{klos}.

\glsfirst{svdd} was proposed for data domain description, which means it is targeted to classify the input as belonging to the \gls{dataset} or not (known or unknown).
In general, any one-class or binary classifier can be applied in such cascade approach to reject or accept the test sample as belonging to one of the known classes and further defining which class it is.
It is similar to the framework proposed by \citet{Cortes2016} in which a rejection function is trained along with a classifier, however, in open-set scenario the classifier should be multiclass.
For the case in which the rejection function accepts a sample, any multiclass classifier can be applied to choose the correct class.
In case of rejection, samples are classified as unknown.

\citet{Costa2012,Costa2014} proposed the \gls{svmdbc}, an extension of the \gls{svm} classifier aiming at a more restrictive specialization on the positive class of the binary classifier.
For this purpose, they move the hyperplane a value $\epsilon$ towards the positive class (in rare cases backwards).
The value $\epsilon$ is obtained by minimizing the \gls{tde}.
For multiclass classification, the \gls{ova} approach can be employed and \gls{svmdbc} uses the \gls{rbf} kernel.
Despite the specialized approach towards the positive class, \gls{svmdbc} cannot ensure a bounded \gls{plos}.

\citet{Scheirer2013} formalized the open-set recognition problem and proposed an extension upon the \gls{svm} classifier called \gls{svm1vs}.
Similar to the works of \citet{Costa2012,Costa2014}, they move the main hyperplane either direction depending on the \gls{osr}.
In addition, a second hyperplane, parallel to the main one, is created such that the positive class is between the two hyperplanes.
This second hyperplane allows the samples ``behind'' the positive class to be classified as negative.
Then a refinement step is performed on both hyperplanes to balance \gls{osr} and \gls{er}.
According to the authors, the method works better with the linear kernel, as the second plane does not provide much benefit for an \gls{rbf} kernel which has a naturally occurring upper bound.
A \gls{ova} approach is used to combine the binary classifiers for open-set multiclass classification.

\citet{Scheirer2014} proposed the \gls{wsvm}.
The authors proposed the \gls{cap} model which decreases the probability of a test sample to be considered as belonging to one of the known classes when it is far away from the training samples.
They use two stages for classification: a \gls{cap} model based on a one-class classifier followed by a binary classifier with normalization based on \gls{evt}.
The binary classifier seeks to improve discrimination and its normalization has two steps.
The first aims at obtaining the probability of a test sample to belong to a positive/known class and the second step estimates the probability of it not being from the negative classes.
Product of both probabilities is the final probability of the test sample to belong to a positive/known class.
\gls{wsvm} uses the \gls{rbf} kernel and also the \gls{ova} approach and ensures a bounded \gls{klos} due to its one-class model.

\citet{Jain2014} proposed the \gls{pisvm}, also based on the \gls{evt}.
It is an algorithm for estimating the unnormalized posterior probability of class inclusion.
For each known class, a Weibull distribution \cite{Coles2001} is estimated based on the smallest decision values of the positive training samples.
The binary classifier for each class is an \gls{svm} with \gls{rbf} kernel trained using the \gls{ova} approach, i.e., the samples of all remaining classes are considered as negative samples.
They introduce the idea of \gls{ccv} which is similar to the open-set \gls{gs} formalized by \citet{MendesJunior2019}.
For a test sample, \gls{pisvm} chooses the class for which the decision value produces the maximum probability of inclusion.
If that maximum is below a given threshold, the input is marked as unknown.
\gls{pisvm} is not able to ensure a bounded \gls{plos} and, consequently, nor the \gls{klos}.

Also applying \gls{evt}, \citet{Rudd2018} proposed a method purely based on Weibull extreme value distributions, named the \gls{evm}.
Their method generates a distribution for each of the known instances aiming at a separation from instances of other classes.
In a later step, those distributions are summarized aiming at a tight probabilistic representation for each of the known classes.
Test phase comprises the verification of pertinence to each of those distributions and examples are rejected when they are outliers for every distribution.
As \gls{evm} creates \gls{cap} models, it is able to bound the \gls{klos}.

Applying \gls{evt} in open-set recognition has been a recent research focus.
\citet{Scheirer2017} presents an overview of how \gls{evt} have been recently applied to visual recognition, mainly on the context of open-set recognition.
Notice that some of the previous \gls{evt}-based works are not capable of ensuring a bounded \gls{klos} by solely relying on \gls{evt} models.
In the case of \citet{Scheirer2014}, they ensure bounded \gls{klos} based on one-class models but not on \gls{evt} models.
For \gls{pisvm} \cite{Jain2014}, the authors did not prove their method is able to bound \gls{klos}.
In fact, \gls{pisvm} can leave an unbounded \gls{klos} when the value of the bias term of the \gls{svm} model is in the range of scores used to fit the Weibull model.
As \gls{evt} for open-set recognition is a growing research area, we highlight that one advantage of our proposed method, compared to \gls{evt}-based ones, relies on its simplicity.
The proposed method is defined purely as a convex optimization problem, as it is for \gls{svm}.
\gls{evt}-based methods require post-processing after an \gls{svm} model is trained, or---as it is the case of \gls{evm}---it requires an expensive model calculated per training instance.
At testing phase, our proposed method requires just the prediction from the obtained \gls{svm} model, while \gls{svm}- and \gls{evt}-based methods require extra predictions from \gls{evt} models for each binary classifier that composes the multiclass one.

\citet{MendesJunior2017} have shown that thresholding ratio of distances in the feature space for a nearest neighbor classifier is more accurate at predicting unknown samples in an open-set problem.
The effectiveness of working with ratio of decision scores is confirmed by \citet{Vareto2017}, in which one of the best rejection threshold estimated for a face-recognition problem is established based on the ratio of the two highest scores obtained by a voting from a set of binary classifiers.

It is worth noticing that well-known machine learning areas have been investigated from the point of view of the open-set scenario, e.g., domain adaptation \cite{Busto2017,Saito2018,Fu2019,Liu2019b}, genetic programming~\cite{Neira2018}, object detection~\cite{Dhamija2020}, incremental learning~\cite{Dang2019}, etc., taking into account particularities from open-set recognition.
Beyond those methods specifically proposed for open-set setups, many other solutions in literature can be investigated to be extended for open-set scenarios.
In general, any binary classification method that aims at decreasing false positive rate \cite{Moraes2016} would potentially recognize unknown samples when composing such classifiers with \gls{ova} approach.

In general, recent solutions for open-set recognition problems have focused on methods that use samples of all known classes for training models for individual classes.
That is different from generative approaches that checks if the test sample is in the distribution of each of the known classes, as it tries to use data from all classes for generating the model for each class.
It differs from \gls{ro} \cite{Tax2008,Bartlett2008} in the sense that we do not want just to postpone decision making.
Moreover, open-set recognition differs from domain adaptation and from transfer learning in the sense that transferring knowledge from one domain to another does not ensure the ability of identifying samples belonging to unknown classes.

Also, multiple researches have been recently accomplished considering open-set recognition for multiple applications, e.g., audio recognition~\cite{Saki2019}, user identity verification~\cite{Tornai2019}, camera model identification~\cite{MendesJunior2019}, human activity recognition~\cite{Yang2019a}, etc.
\citet{Geng2020} present a survey of the literature on open-set recognition; we refer the reader to their work for a more complete review of the literature.

\section{\texorpdfstring{\glsdesc{ossvm}}{OSSVM}}\label{sec:ssvm}

One cannot ensure that the positive class of the traditional \gls{svm} has a bounded \gls{plos}, even when the \gls{rbf} kernel is used.
The main characteristic of the proposed \gls{ossvm} is that a high enough regularization parameter ensures a bounded \gls{plos} for every known class of interest, consequently, a bounded \gls{klos} in \gls{ova} multiclass level.
The regularization parameter is a weight for optimizing the \gls{ru} in relation to the \gls{er} measured on training data.
In Section \ref{sec:ssvm-bounding-open-space}, we present how to ensure a bounded \gls{plos} using \gls{rbf} kernel.
In Section \ref{sec:ssvm-choosing-bias-term}, we present the formulation of the optimization problem of the \gls{ossvm}.
To begin with, in Section \ref{sec:svm-basics}, we present some basic aspects of the \gls{svm} classifier.

\subsection{Basic aspects of \texorpdfstring{\glsdesc{svm}}{SVM}}\label{sec:svm-basics}

\gls{svm} is a binary classifier that, given a set $X$ of training samples $\mathbf{x}_i \in \mathbb{R}^d$ and the corresponding labels $y_i \in \{-1, 1\}$, $i = 1, \dots, m$, it finds a maximum-margin hyperplane that separates $\mathbf{x}_i$ for which $y_{i} = -1$ from $\mathbf{x}_j$ for which $y_{j} = 1$ \cite{Cortes1995}.
We consider the soft margin case with parameter $C$.

The primal optimization problem is usually defined as
\begin{equation*}
  \min_{\mathbf{w},b,\xi} \frac{1}{2}\|\mathbf{w}\|^{2} + C\sum_{i=1}^{m}\xi_{i},
\end{equation*}
\begin{align}
  \label{eq:constraint-1}
  \mbox{s.t.}\ & y_i(\mathbf{w}^{T}\mathbf{x}_{i} + b) \ge 1 - \xi_{i}, \ \forall i,\\
  \label{eq:constraint-2}
               & \xi_{i} \ge 0, \ \forall i.
\end{align}

To solve this optimization problem, we use the Lagrangian method to create the dual optimization problem.
In this case, the final Lagrangian is defined as
\begin{equation}
  \label{eq:lagrangian}
  \mathcal{L}(\alpha) = \sum_{i=1}^{m}\alpha_{i} - \frac{1}{2}\|\mathbf{w}\|^{2},
\end{equation}
in which $\mathbf{w} = \sum_{i=1}^{m}\alpha_{i}y_{i}\mathbf{x}_{i}$ and $\alpha_{i} \in \mathbb{R}$, $i = 1, \dots, m$, are the Lagrangian multipliers.
Then, the optimization problem now is defined as
\begin{align}
  \label{eq:lagrangian-inv}
  \min_{\alpha}\ & \mathcal{W}(\alpha) = -\mathcal{L}(\alpha) = \frac{1}{2}\|\mathbf{w}\|^{2} - \sum_{i=1}^{m}\alpha_{i},\\
  \label{eq:slack-constraint}
  \mbox{s.t.}\ & 0 \le \alpha_{i} \le C, \ \forall i,\\
  \label{eq:sum_0_constraint}
                 & \sum_{i=1}^{m}\alpha_{i}y_{i} = 0.
\end{align}

The decision function of a test sample $\mathbf{x}$ comes from the constraint in Equation~\eqref{eq:constraint-1} and is defined as
\begin{equation}
  \label{eq:decision-function}
  f(\mathbf{x}) = \sign(\mathbf{w}^T \mathbf{x} + b) = \sign\left(\sum_{i=1}^{m} y_i \alpha_i \mathbf{x}_i^T \mathbf{x} + b\right).
\end{equation}

\citet{Boser1992} proposed a modification in \gls{svm} for the cases in which the training data are not linearly separated in the feature space.
Instead of linearly separating the samples in the original space $\mathcal{X}$ of the training samples in $X$, the samples are projected onto a higher dimensional space $\mathcal{Z}$ in which they are linearly separated.
This projection is accomplished using the kernel trick \cite{Mercer1909}.
One advantage of this method is that in addition to separating non-linear data, the optimization problem of the \gls{svm} remains almost the same: instead of calculating the inner product $\mathbf{x}^T\mathbf{x}'$, it uses a kernel $K(\mathbf{x}, \mathbf{x}')$ that is equivalent to the inner product $\phi(\mathbf{x})^T\phi(\mathbf{x}')$ in a higher dimensional space $\mathcal{Z}$, in which $\phi: \mathcal{X} \mapsto \mathcal{Z}$ is a projection function.
When using the kernel trick, we do not need to know the $\mathcal{Z}$ space explicitly.

Using kernels, the decision function of a test sample $\mathbf{x}$ becomes
\begin{equation}
  \label{eq:decision-function-kernel}
  f(\mathbf{x}) = \sign\left(\sum_{i=1}^{m} y_i \alpha_i K(\mathbf{x}_i, \mathbf{x}) + b\right).
\end{equation}

The most used kernel for \gls{svm} is the \gls{rbf} kernel \cite{Scholkopf2001}, defined as follows.
\begin{equation}
  \label{eq:kernel-rbf}
  K(\mathbf{x}, \mathbf{x}') = e^{-\gamma \|\mathbf{x} - \mathbf{x}'\|^2}.
\end{equation}
It is proved that using this kernel, the projection space $\mathcal{Z}$ is an $\infty$-dimensional space \cite{Scholkopf2001}.

\subsection{Ensuring a bounded \texorpdfstring{\gls{plos}}{PLOS}}\label{sec:ssvm-bounding-open-space}

By simply employing the \gls{rbf} kernel, we cannot ensure the \gls{plos} is bounded.

\begin{theorem}
  \label{thm:unbounded-positive}
  \gls{svm} with \gls{rbf} kernel has a bounded \glsfirst{plos} if and only if the bias term $b$ is negative.\footnote{%
    In some implementations, including the \gls{libsvm} library \cite{Chang2011}, the decision function is defined as $f(\mathbf{x}) = \sign(\mathbf{w}^T \mathbf{x} - \rho)$ instead of the one in Equation~\eqref{eq:decision-function}.
    In that case, instead of ensuring a negative bias term $b$, one must ensure a positive bias term $\rho$ to bound the \gls{plos}.%
  }
  \begin{proof}
    We know that
    \begin{equation}
      \label{eq:limit}
      \lim_{d \rightarrow \infty} K(\mathbf{x}, \mathbf{x}') = 0,
    \end{equation}
    in which $K(\mathbf{x}, \mathbf{x}')$ is the \gls{rbf} kernel and $d = \|\mathbf{x} - \mathbf{x}'\|$.
    For the cases in which a test sample $\mathbf{x}$ is far away from every support vector $\mathbf{x}_i$, we have that
    \begin{equation*}
      \sum_{i=1}^{m} y_i \alpha_i K(\mathbf{x}_i, \mathbf{x})
    \end{equation*}
    also tends to 0.
    From Equation~\eqref{eq:decision-function-kernel} it follows that
    \begin{equation*}
      f(\mathbf{x}) = \sign\left(b\right)
    \end{equation*}
    when $\mathbf{x}$ is far away from the support vectors.
    Therefore, for negative values of $b$, $f(\mathbf{x})$ is always negative for far away $\mathbf{x}$ samples.
    That is, samples in a bounded region of the feature space will be classified as positive.
    For the \emph{only if} direction, let $b$ be positive.
    Then, for far away $\mathbf{x}$ examples we have $f(\mathbf{x}) = \sign(b) > 0$, i.e., positively classified samples will be in an unbounded region of the feature space when $b$ is positive.
  \end{proof}
\end{theorem}

Theorem \ref{thm:unbounded-positive} can be applied not only to the \gls{rbf} kernel of Equation~\eqref{eq:kernel-rbf} but to any radial basis function \cite{Buhmann2003} kernel satisfying Equation~\eqref{eq:limit}, e.g., \glsdesc{tst} kernel, \glsdesc{rq} kernel, and \glsdesc{imq} kernel~\cite{Souza2010}, however, for the remaining part of this work, we focus on the \gls{rbf} kernel of Equation~\eqref{eq:kernel-rbf}.

Figure~\ref{fig:figPlottingkernel} depicts the rationale behind Theorem~\ref{thm:unbounded-positive} on a 2-dimensional synthetic dataset.
The $z$ axis represents the decision values for which possible \gls{2d} test samples $(x, y)$ would have for different regions of the feature space.
Training samples are normalized between 0 and 1.
Note in the subfigures that for possible test samples far away from the training ones, e.g., $(2, 2)$, the decision value approaches the bias term $b$.
Note in Figure~\ref{fig:figPlottingkernel3d__boat_forboundaries__cls03__01__gammaexp+04} that an unbounded region of the feature space would have samples classified as positive.
Consequently, all those samples would be classified as class 3 by the final multiclass-from-binary classifier.
In general \gls{svm} usage, both positive and negatives biases occur as $b$ depends on the training data.

In case of \gls{svm}s without explicit bias term \cite{Vogt2002,Kecman2005}, $b=0$ is implicit.\footnote{%
  The main difference from the \gls{svm} without bias term to the traditional \gls{svm} is that the constraint in Equation \eqref{eq:sum_0_constraint} does not exist in the dual formulation.%
}
Consequently, the decision function is defined as
\begin{equation*}
  f(\mathbf{x}) = \sign\left(\sum_{i=1}^{m} y_i \alpha_i K(\mathbf{x}_i, \mathbf{x})\right).
\end{equation*}
For test samples far away from the support vectors, we have that $\sum_{i=1}^{m} y_i \alpha_i K(\mathbf{x}_i, \mathbf{x})$ converges to 0 from the bottom or from the above, depending on the training samples.
Consequently, a bounded \gls{plos} cannot be ensured in all cases.

\figPlottingkernel{}

Theorem~\ref{thm:unbounded-positive} also provides a solution to the problem of unbounded \gls{plos}.
We can ensure a bounded \gls{plos} by simply employing an \gls{rbf} kernel and ensuring a negative $b$.
In Section~\ref{sec:ssvm-choosing-bias-term}, we present a new \gls{svm} optimization objective that optimizes the margin while ensuring the bias term $b$ is negative.

\subsection{Optimization to ensure negative bias term \texorpdfstring{$b$}{b}}\label{sec:ssvm-choosing-bias-term}

As we discussed in Section~\ref{sec:ssvm-bounding-open-space}, we must ensure a negative $b$ to obtain a bounded \gls{plos}.
For this, we define the \gls{ossvm} optimization problem as
\begin{equation}
  \label{eq:ssvm-primal-opt}
  \min_{\mathbf{w},b,\xi} \frac{1}{2}\|\mathbf{w}\|^{2} + C\sum_{i=1}^{m}\xi_{i} + \lambda b,
\end{equation}
subject to the same constraints defined in Equations~\eqref{eq:constraint-1} and \eqref{eq:constraint-2}, in which $\lambda$ is a regularization parameter that trades off between the \gls{er} and the \gls{ru}.

From Equation~\eqref{eq:ssvm-primal-opt}, the dual formulation has the same Lagrangian defined in Equation~\eqref{eq:lagrangian}.
Consequently, we have to optimize the same function as defined in Equation~\eqref{eq:lagrangian-inv} with the constraint in Equation~\eqref{eq:slack-constraint}.
However, the constraint in Equation~\eqref{eq:sum_0_constraint} is replaced by the constraint
\begin{equation}
  \label{eq:sum_lambda_constraint}
  \sum_{i=1}^{m}\alpha_{i}y_{i} = \lambda.
\end{equation}

The same \gls{smo} algorithm proposed by \citet{Platt1998}, with the \gls{wss} proposed by \citet{Fan2005}, for optimizing ensuring the constraint in Equation~\eqref{eq:sum_0_constraint} can be applied to this optimization containing the constraint of the Equation \eqref{eq:sum_lambda_constraint}.
As the main idea of the \gls{smo} algorithm is to ensure that $\sum\alpha_{i}y_{i}$ remains the same from one iteration to the other, before the optimization starts, we initialize $\alpha_{i}$ such that $\sum\alpha_{i}y_{i} = \lambda$.
For this, we let $\alpha_{i} = \lambda/m_{p}$, $\forall i$ such that $y_{i} = 1$, in which $m_{p}$ is the number of positive training samples.

\begin{proposition}
  \label{prop:maximum-lambda}
  For the \gls{svm} with soft margin, the maximum valid value for $\lambda$ is $C m_{p}$.
  \begin{proof}
    From Equation~\eqref{eq:slack-constraint}, $0\le\alpha_{i}\le C$.
    The maximum value $\lambda=\sum\alpha_{i}y_{i}$ is thus obtained by setting $\alpha_{i} = C$ for $i$ such that $y_{i} = 1$ and setting $\alpha_{i} = 0$ for $i$ such that $y_{i} = -1$.
    This yields $\lambda \le C m_{p}$
  \end{proof}
\end{proposition}

During optimization, we must ensure \mbox{$\lambda \le C m_{p}$} given that if \mbox{$\lambda > C m_{p}$}, the constraint in Equation~\eqref{eq:slack-constraint} would be broken for some $\alpha_{i}$.

Despite Proposition~\ref{prop:maximum-lambda} saying that it is allowed \mbox{$\lambda = C m_{p}$}, when it happens, we have that $\alpha_{i} = C$ for $y_{i} = 1$ and $\alpha_{i} = 0$ for $y_{i} = -1$, and there will be no optimization.
In this case, despite satisfying the constraints, there is no flexibility for changing values of $\alpha_{i}$ because, for each pair $\alpha_{i}, \alpha_{j}$ selected by the \gls{wss} algorithm, we must update $\alpha_{i} = \alpha_{i} + \nabla_{\alpha}$, $\alpha_{j} = \alpha_{j} + \nabla_{\alpha}$ when $y_{i} \ne y_{j}$ and $\alpha_{i} = \alpha_{i} - \nabla_{\alpha}$, $\alpha_{j} = \alpha_{j} + \nabla_{\alpha}$ when $y_{i} = y_{j}$.
For any $\nabla_{\alpha} \ne 0$, the constraint $0 < \alpha_{i} < C$ would break for either $\alpha_{i}$ or $\alpha_{j}$, for any selected pair.
Then, in practice, we \gls{gs} $\lambda$ in the interval $0 \le \lambda < C m_{p}$.

Proposition~\ref{prop:maximum-lambda} holds true for any kernel, however, the formulation in Equation~\eqref{eq:ssvm-primal-opt} only has the open-set properties previously discussed with an \gls{rbf} kernel, as we have observed through Theorem~\ref{thm:unbounded-positive}.

\begin{proposition}
  \label{prop:negative-bias-term}
  There exists some $\lambda$ such that we can obtain a bias term $b < 0$ for the \gls{ossvm} with an \gls{rbf} kernel $K$ such that $0 < K(\mathbf{x}, \mathbf{x}') \le 1$ when $C \ge 1$.
  \begin{proof}
    From the \gls{kkt} conditions, the bias term is defined as
    \begin{align*}
      b &= y_{i} - \sum_{j=1}^{m}\alpha_{j}y_{j}K\left(\mathbf{x}_{i}, \mathbf{x}_{j}\right)\\
        &= y_{i} - \sum_{\substack{j=1:\\y_{j} = 1}}^{m}\alpha_{j}K\left(\mathbf{x}_{i}, \mathbf{x}_{j}\right) + \sum_{\substack{j=1:\\y_{j} = -1}}^{m}\alpha_{j}K\left(\mathbf{x}_{i}, \mathbf{x}_{j}\right),
    \end{align*}
    for any $i$ such that $0 < \alpha_{i} < C$.
    Now, let us consider two possible cases: (1) $y_{i} = 1$ and (2) $y_{i} = -1$.
    For \textbf{Case (1)}, we have
    \begin{equation*}
      b = 1 - \alpha_{i} - \sum_{\substack{j=1:\\y_{j} = 1,\\j\ne i}}^{m}\alpha_{j}K\left(\mathbf{x}_{i}, \mathbf{x}_{j}\right) + \sum_{\substack{j=1:\\y_{j} = -1}}^{m}\alpha_{j}K\left(\mathbf{x}_{i}, \mathbf{x}_{j}\right),
    \end{equation*}
    as $K\left(\mathbf{x}_{i}, \mathbf{x}_{i}\right) = 1$.
    Note that $0 < K(\mathbf{x}, \mathbf{x}') \le 1$.
    To show that there exists some $\lambda$ such that $b < 0$, we analyze the worst case, i.e., when the kernel in the second summation---for negative training samples---is $1$.
    Then, we have
    \begin{equation*}
      b = 1 - \alpha_{i} - \sum_{\substack{j=1:\\y_{j} = 1,\\j\ne i}}^{m}\alpha_{j}K\left(\mathbf{x}_{i}, \mathbf{x}_{j}\right) + \sum_{\substack{j=1:\\y_{j} = -1}}^{m}\alpha_{j}.
    \end{equation*}
    From Equation \eqref{eq:sum_lambda_constraint}, we have
    \begin{equation}
      \label{eq:sum_lambda_constraint_neg_equality}
      \sum_{\substack{j=1:\\y_{j} = -1}}^{m}\alpha_{j} = \sum_{\substack{j=1:\\y_{j} = 1}}^{m}\alpha_{j} - \lambda,
    \end{equation}
    then
    \begin{equation*}
      b = 1 - \sum_{\substack{j=1:\\y_{j} = 1,\\j\ne i}}^{m}\alpha_{j}K\left(\mathbf{x}_{i}, \mathbf{x}_{j}\right) + \sum_{\substack{j=1:\\y_{j} = 1\\j\ne i}}^{m}\alpha_{j} - \lambda
    \end{equation*}
    Analyzing the worst case again, considering $\alpha_{j} = C$ for positive training samples, with $j \ne i$, we have
    \begin{align*}
      b &= 1 - C\sum_{\substack{j=1:\\y_{j} = 1,\\j\ne i}}^{m}K\left(\mathbf{x}_{i}, \mathbf{x}_{j}\right) + C\left(m_{p} - 1\right) - \lambda\\
        &= 1 + C m_{p} - C - C\sum_{\substack{j=1:\\y_{j} = 1,\\j\ne i}}^{m}K\left(\mathbf{x}_{i}, \mathbf{x}_{j}\right) - \lambda.
    \end{align*}
    To ensure $b < 0$ it is sufficient to let
    \begin{equation}
      \label{eq:lambda_min_pos}
      \lambda > 1 + C m_{p} - C - C\sum_{\substack{j=1:\\y_{j} = 1,\\j\ne i}}^{m}K\left(\mathbf{x}_{i}, \mathbf{x}_{j}\right).
    \end{equation}
    From Proposition~\ref{prop:maximum-lambda}, we have that $\lambda \le C m_{p}$.
    This fact, along with Equation~\eqref{eq:lambda_min_pos}, leads to the constraint
    \begin{equation*}
      1 + C m_{p} - C - C\sum_{\substack{j=1:\\y_{j} = 1,\\j\ne i}}^{m}K\left(\mathbf{x}_{i}, \mathbf{x}_{j}\right) < C m_{p},
    \end{equation*}
    which simplifies to
    \begin{equation*}
      1 - C - C\sum_{\substack{j=1:\\y_{j} = 1,\\j\ne i}}^{m}K\left(\mathbf{x}_{i}, \mathbf{x}_{j}\right) < 0.
    \end{equation*}
    Analyzing the worst case one more time, considering $K\left(\mathbf{x}_{i}, \mathbf{x}_{j}\right) \approx 0$ for the positive training samples other than $\mathbf{x}_{i}$, and replacing the term with kernel summation to $\epsilon$, we have
    \begin{equation*}
      1 - C - \epsilon < 0.
    \end{equation*}
    As $\epsilon > 0$, it is always possible to satisfy this constraint for $C \ge 1$.

    For \textbf{Case (2)}, we have
    \begin{equation*}
      b = - 1 - \sum_{\substack{j=1:\\y_{j} = 1}}^{m}\alpha_{j}K\left(\mathbf{x}_{i}, \mathbf{x}_{j}\right) + \sum_{\substack{j=1:\\y_{j} = -1}}^{m}\alpha_{j}K\left(\mathbf{x}_{i}, \mathbf{x}_{j}\right).
    \end{equation*}
    Considering the worst case for the values of the kernel for negative samples and using the equality in Equation \eqref{eq:sum_lambda_constraint_neg_equality}, we have
    \begin{equation*}
      b = -1 - \sum_{\substack{j=1:\\y_{j} = 1}}^{m}\alpha_{j}K\left(\mathbf{x}_{i}, \mathbf{x}_{j}\right) + \sum_{\substack{j=1:\\y_{j} = 1}}^{m}\alpha_{j} -  \lambda.
    \end{equation*}
    Considering the highest possible value for $b$, by setting $\alpha_{j} = C$ for positive samples, we have
    \begin{equation*}
      b = -1 - C\sum_{\substack{j=1:\\y_{j} = 1}}^{m}K\left(\mathbf{x}_{i}, \mathbf{x}_{j}\right) + C m_{p} - \lambda.
    \end{equation*}
    In this case, to ensure $b < 0$ it is sufficient to let
    \begin{equation*}
      \lambda > C m_{p} - 1 - C\sum_{\substack{j=1:\\y_{j} = 1}}^{m}K\left(\mathbf{x}_{i}, \mathbf{x}_{j}\right),
    \end{equation*}
    which is possible to obtain for any value of $C$ in such way that $\lambda \le C m_{p}$ of Proposition~\ref{prop:maximum-lambda} is also satisfied.
  \end{proof}
\end{proposition}

In Proposition~\ref{prop:negative-bias-term}, we considered a very extreme case for the proof.
For example, in Case (1)---for $i$ such that $y_{i} = 1$---we considered $K\left(\mathbf{x}_{i}, \mathbf{x}_{j}\right) = 1$ for $j$ such that $y_{j} = -1$ and $K\left(\mathbf{x}_{i}, \mathbf{x}_{j}\right) \approx 0$ for $j$ such that $y_{j} = 1, j \ne i$.
It means that all negative samples have the same feature vector of sample $\mathbf{x}_{i}$ under consideration and all positive samples are far away from sample $\mathbf{x}_{i}$.
In practice, we do not have the $\lambda$ nearly as constrained as in the proof to ensure a negative bias term.
Moreover, in our experiments with the \gls{mcsvm}, we observed that oftentimes the bias term is negative for a binary classifier trained with the \gls{ova} approach, i.e., it is often the case that even with $\lambda=0$ the bias will be negative.
More details about this behavior is shown in Section \ref{sec:remarks}.

One can argue that instead of defining the new optimization problem of Equation~\eqref{eq:ssvm-primal-opt}, the \gls{svm} decision function of Equation~\eqref{eq:decision-function-kernel} could be simply changed to
\begin{equation}
  \label{eq:decision-function-with-epsilon}
  f(\mathbf{x}) = \sum_{i=1}^{m} y_i \alpha_i K(\mathbf{x}_i, \mathbf{x}) + b > \epsilon
\end{equation}
in order to bound the \gls{plos}, in which $\epsilon$ is a parameter that could be defined after the optimization process.
As Equation~\eqref{eq:decision-function-with-epsilon} is equivalent to
\begin{equation*}
  f(\mathbf{x}) = \sign\left(\sum_{i=1}^{m} y_i \alpha_i K(\mathbf{x}_i, \mathbf{x}) + b - \epsilon\right),
\end{equation*}
the \gls{plos} could be bounded if and only if $\epsilon \ge b$.
However, this simplified approach has drawbacks.
The value of $b$ can only be known after the \gls{svm} optimization process is completed and it depends on the training data.
Notice that the introduction of $\epsilon$ is equivalent to performing a parallel translation of the optimal hyperplane.
As \gls{svm} only optimizes the \gls{er}, the final model would not be optimal neither according to the \gls{er} itself nor the \gls{osr}.
On the other hand, \gls{ossvm} optimizes both the \gls{er} and the \gls{osr} and the separation hyperplane obtained by \gls{ossvm} is not necessarily parallel to the position of the hyperplane that would be obtained by the traditional \gls{svm} on the same training data.

In Appendix \ref{appendix:sec:ssvm-formulation}, we present the complete formulation of the optimization problem for the \gls{ossvm} classifier.\footnote{%
  Source code, extended upon the \gls{libsvm} implementation \cite{Chang2011}, is available through \url{https://pedrormjunior.github.io/OSSVM.html}.%
}

\paragraph*{Choosing the $\lambda$ parameter for the \gls{ossvm}}
Proposition~\ref{prop:negative-bias-term} states that we can find a $\lambda$ parameter that ensures a bounded \gls{plos} for the optimization problem presented above.
To ensure this, models with a non-negative bias term receive accuracy of $-\infty$ on the validation set, during the \gls{gs}.
Nevertheless, we cannot ignore that, in special circumstances, certain $\lambda$ values allow a negative bias term during the \gls{gs} but not for training in the whole set of training samples.
In this case, once the parameters are obtained by \gls{gs}, if the obtained $\lambda$ does not ensure a negative bias term for the whole training set, one would need to retrain the classifier with an increased value for $\lambda$, until a negative bias term is obtained for the final model.
However, for grid search, we assume the distribution of the validation set, a subset of the training set, represents the distribution of the training set; that is one possible explanation as for why in our experiments we did not need to retrain the classifier with a value of $\lambda$ larger than the one obtained during grid search, as all values of $\lambda$ obtained during \gls{gs} were able to ensure a negative bias term for all binary classifiers.

In summary, \gls{ossvm} optimizes the bias term $b$ in order to ensure a bounded \gls{plos} for every binary classifier.
The \gls{plos} is bounded if and only if $b < 0$.
The $\lambda$ parameter introduced by \gls{ossvm} is responsible for optimizing $b$, taking into account the \gls{ru}.

\section{Experiments}
\label{sec:experiments}

In this section, we present the experiments and details for comparing the proposed method with the existing ones in the literature, discussed in Section \ref{sec:related-work}.
In Section \ref{sec:experiments-baselines-summary}, we summarize the baselines.
In Section \ref{sec:evaluation-measures}, we describe the evaluation measures used in our experiments.
In Section \ref{sec:datasets}, we describe the \glspl{dataset} and features.
In Section \ref{sec:results-decision-boundaries}, we present initial results regarding the behavior of the methods in feature space.
Finally, we present the results with statistical tests in Section \ref{sec:results}, finishing this section with some remarks in Section \ref{sec:remarks}.

\subsection{Baselines}
\label{sec:experiments-baselines-summary}

In this work, we employ the one-class--based method of \citet{Pritsos2013} for comparison, hereinafter referred to as \gls{mcocsvm}.
Although \citet{Pritsos2013} use a \gls{ocsvm} as the method to define a rejection function for each class, one could also use a \gls{svdd}, since it is also a form of one-class classifier.
We also implemented this alternative to \gls{mcocsvm} and refer to it as \gls{mcsvdd}.

Despite dealing with a multiclass problem, \citet{Costa2012,Costa2014} evaluated their method \gls{svmdbc} in the binary fashion by obtaining the accuracy of individual binary classifiers.
They did not present the multiclass version of the classifier directly.
Therefore, in this work, we consider their method with the \gls{ova} approach in the experiments.
The test sample is classified as unknown when no binary classifier classifies it as positive.
Complementarily, it is classified as the most confident class when one or more classifiers tags it as positive.

Besides those methods, we have employed \gls{mcsvm} \cite{Chang2011}, \gls{svm1vs} \cite{Scheirer2013}, \gls{wsvm} \cite{Scheirer2014}, \gls{pisvm} \cite{Jain2014}, and \gls{evm} \cite{Rudd2018} as baselines for our experiments.
Among those methods, only one-class--based methods, \gls{evm} and \gls{ossvm} are able to ensure a bounded \gls{klos}.
We summarize the methods in Table \ref{tab:baselines-summary}.

\tableBaselinesSummary{}

We have employed the open-set \gls{gs} approach for the existing methods in the literature, as in previous work it has been demonstrated to better estimate the parameters for classifiers in open-set scenarios \cite{MendesJunior2019}.
Except for \gls{evm}, all compared methods are \gls{svm}-based.
Parameters of the proposed method and baselines were obtained through grid search.
All \gls{svm}-based methods have fixed $C=1$.
For \gls{rbf}-based methods, the $\gamma$ parameter were searched in $\{2^{-15}, 2^{-13}, \dots, 2^{15}\}$.
The $\nu$ parameter of \gls{mcocsvm} were searched among 21 values linearly spaced in $[0, 1]$.
The $p_{A}$ and $p_{\Omega}$ parameters of the \gls{svm1vs} were both searched in $\{2^{-3}, 2^{-2}, \dots, 2^{2}\}$.
The $\delta_{\tau}$ threshold for \gls{wsvm}'s CAP model were fixed in 0.001, as specified by the authors~\cite{Scheirer2014}, and the $\delta_{R}$ were searched in $\{2^{-7}, 2^{-6}, \dots, 2^{0}\}$.
The \gls{pisvm}'s threshold were searched among 20 values linearly spaced in $[0, 1)$.
The $\lambda$ parameter of \gls{ossvm} were searched among 20 values linearly spaced in $[0, Cm_{p})$.

\subsection{Evaluation measures}\label{sec:evaluation-measures}

Most of the proposed evaluation performance measures in the literature are focused on binary classification, e.g., traditional classification accuracy, \gls{fmeasure}, etc.\@ \cite{Sokolova2009}.
Even the ones proposed for multiclass scenarios---e.g., average accuracy, multiclass \gls{fmeasure}, etc.\@---usually consider only the closed-set scenario.
Recently, \citet{MendesJunior2017} have proposed \gls{na} and \gls{osfm} for multiclass open-set recognition problems.
In this work, we apply such measures and further extend the \gls{na} to \gls{hna}, based on the harmonic mean \cite{Mitchell2004}, as shown in Equation~\eqref{eq:hna}.

\begin{equation}
  \label{eq:hna}
  \glssymbol{hna} =
  \begin{cases}
    0, \text{if } \glssymbol{aks} = 0 \text{ or } \glssymbol{aus} = 0,\\
    \displaystyle\frac{2}{\displaystyle\left(\frac{1}{\displaystyle\glssymbol{aks}} + \frac{1}{\displaystyle\glssymbol{aus}}\right)}, \text{otherwise}.
  \end{cases}
\end{equation}
In this case, \glstext{aks} is the \glsdesc{aks} and \glstext{aus} is the \glsdesc{aus}.
\Gls{aks} is the accuracy obtained on the testing instances that belong to one of the classes with which the classifier was trained.
\Gls{aus} is the accuracy on the testing instances whose classes have no representative instances in the training set.

One advantage of \gls{hna} over \gls{na} is that when a classifier performs poorly on \gls{aks} or \gls{aus}, \gls{hna} drops toward 0.
One biased classifier that blindly classifies every example as unknown would receive \gls{na} of 0.5 while \gls{hna} would be 0.
However, notice that 0.5 is not the worst possible accuracy for \gls{na}, as some methods---trying to correctly predict test labels---can have its \gls{na} smaller than 0.5.
The worst case for \gls{na}---when it is 0---would be when all known samples are classified as unknown and all unknown samples classified as belonging to one of the known classes.
On the other hand, the worst case for \gls{hna} would be when at least one of such cases happens.

For experiments in this work, we have considered \gls{na}, \gls{hna}, \gls{osfmM}, and \gls{osfmm}.
For a fair comparison with previous methods in the literature \cite{Scheirer2013,Scheirer2014,Jain2014,Rudd2018}---which only showed performance figures using the traditional \gls{fmeasure}---we also present results regarding the traditional multiclass \gls{fmeasure} \cite{Sokolova2009} considering both \gls{fmM} and \gls{fmm}.

\subsection{\texorpdfstring{\Glspl{dataset}}{Datasets}}
\label{sec:datasets}

For validating the proposed method and comparing it with existing methods, we consider \gls{number-of-datasets} \glspl{dataset}.
In the \gls{15-scenes} \gls{dataset} (with 15 classes), the images are represented by a bag-of-visual-word vector created with soft assignment \cite{vanGemert2010} and max pooling \cite{Boureau2010}, based on a codebook of 1000 Scale Invariant Feature Transform (SIFT) codewords \cite{Lowe2004}.
\gls{letter} \gls{dataset} (with 26 classes) represents the letters of the English alphabet (black-and-white rectangular pixel displays).
The \gls{kddcup} \gls{dataset} (with 32 classes\footnote{Aiming at keeping the same setup across all \glspl{dataset} (see Section~\ref{sec:results}), for \gls{kddcup}, we have joined training and testing \glspl{dataset} for the experiments.  As \gls{wsvm} cannot fit the model with classes with few samples, aiming at a paired experiment, we have kept only the classes with 10 or more samples.}) represents an intrusion detection problem on a military network environment and its feature vectors combine continuous and symbolic features.
In the \gls{auslan} \gls{dataset} (with 95 classes), the data was acquired using two Fifth Dimension Technologies (5DT) gloves hardware and two Ascension Flock-of-Birds magnetic position trackers.
In the \gls{caltech-256} \gls{dataset} (with 256 classes), the feature vectors consider a bag-of-visual-words characterization approach, with features acquired with dense sampling, SIFT descriptor for the points of interest, hard assignment \cite{vanGemert2010}, and average pooling \cite{Boureau2010}.
Finally, for the \gls{aloi} \gls{dataset} (with 1000 classes), the features were extracted with the Border/Interior (BIC) descriptor \cite{Stehling2002}.
These \glspl{dataset} or other \glspl{dataset} could be used with different characterizations, however, in this work, we focus on the learning part of the problem rather than on the feature characterization one.

\tableDatasetsInfo{}

For the \gls{imagenet} \gls{dataset}, we performed experiments on 360 classes of \gls{imagenet2010} that has no overlap with the 1000 classes in \gls{imagenet2012}.
Those images were made available by \citet{Bendale2016}.
The network used for feature extraction on those 360 classes was trained on \gls{imagenet2012}.
We used a different \gls{dataset} for training the network aiming at avoiding considering as unknown the classes that could be known from the point view of the network, i.e., classes for which the network learns how to represent.
We trained a \gls{googlenet} network and extracted the features from its last pooling layer.
We applied \gls{pca} to reduce from 1024 features to 100.

For \gls{cifar10} and \gls{mnist} \glspl{dataset}, we employed publicly available networks~\cite[respectively]{TensorflowCIFAR10,TensorflowDeepMNIST} for training and extracting features.
Differently than for \gls{imagenet}, we did not avoid training the networks on the classes to be used on the open-set experiments.
Consequently, each network was trained on all 10 classes included on those \glspl{dataset}.

In Table~\ref{tab:datasets}, we summarize the main features of the considered \glspl{dataset} in terms of number of classes, number of samples, dimensionality, and approximate number of samples per class.
All other datasets in Table~\ref{tab:datasets} not specified herein are obtained from \gls{pmlb} data repository.

\subsection{Decision regions}
\label{sec:results-decision-boundaries}

We start presenting results on artificial \gls{2d} \glspl{dataset} so that \glspl{dr} of each classifier can be visualized.
For these cases, we show the region of the feature space in which a possible test sample would be classified as one of the known classes or unknown.
We also show how each classifier handles the \gls{openspace}.
Figure \ref{fig:cone-torus} depicts the images of \glspl{dr} for the \gls{cone-torus} \gls{dataset} \cite{Kuncheva2004}.
In Figure \ref{subfig:cone-torus__mcossvm_ova_gsic}, as expected, we see that the \gls{ossvm} gracefully bounds the \gls{klos}; any sample that would appear in the white region would be classified as unknown.

\figBoundaries{cone-torus}{\gls{cone-torus}}

\subsection{Results}
\label{sec:results}


We performed a series of experiments simulating an open-set scenario in which 3, 6, 9, and 12 classes are available for training the classifiers.
Remaining classes of each \gls{dataset} are unknown at training phase and only appear on testing stage.
Since different \glspl{dataset} have a different number of known classes, the fraction of unknown classes---or the openness~\citep{Scheirer2013}---varies per \gls{dataset}.
For each \gls{dataset}, method, and number $n$ of \gls{acs}, we run 10 experiments by choosing $n$ random classes for training, among the classes of the \gls{dataset}.
Selected sets of classes are used across each experiment with different classifiers.
In addition, the same samples used for training one classifier $C_i$ is used when training another classifier $C_j$ (a similar setup is adopted for the testing and validation sets), which is referred to as a paired experiment.\footnote{%
  \citet{MendesJunior2019} present additional experiments with the \gls{ossvm} method we are proposing herein.
}

Following the method of \citet{Demsar2006}, we have employed Bonferroni-Dunn statistical test for comparing \gls{ossvm} with baselines considering a confidence interval of 95\%.
In \gls{cd} diagrams in Figure~\ref{fig:CD_normal}, we define \gls{ossvm} as the control method and compare it with baselines for all the measures---\gls{na}, \gls{hna}, \gls{osfmM}, \gls{osfmm}, \gls{fmM}, \gls{fmm}.
We see in those figures that \gls{ossvm} stands at first place for most of the considered measures, however, the \gls{evt}-based methods \gls{pisvm} and \gls{evm} follow it closely.
\gls{ossvm}, then, turns out to be a competitive alternative for recognition in open-set scenarios, with the advantage of its simplicity and guarantee of closing the \gls{klos}.\footnote{%
  Raw results obtained in our experiments as well as the source-code to perform the complete statistical analysis are available through \url{https://pedrormjunior.github.io/OSSVM.html}.%
}


\newcommand\includeCD[2][]{\includegraphics[trim={40pt 15pt 22pt 12pt},clip,width=\linewidth,#1]{#2}} 
\newcommand\subfigureCDwidth{0.495\linewidth}
\newcommand\subfigureCD[3]{%
  \begin{subfigure}{\subfigureCDwidth}
    \centering
    \includeCD{\changedir{CD_diagrams/}CD_#3_#1_O}
    \caption{\Glsdesc{#2}.}
    \label{fig:CD-#2}
  \end{subfigure}%
}

\newcommand\figureCD[2]{
  \begin{figure*}
    \centering
    \subfigureCD{NA}{na}{#1}
    \hfill
    \subfigureCD{HNA}{harmonicNA}{#1}

    \vspace{10pt}

    \subfigureCD{OSFMmacro}{osfmM}{#1}
    \hfill
    \subfigureCD{OSFMmicro}{osfmm}{#1}

    \vspace{10pt}

    \subfigureCD{FMmacro}{fmM}{#1}
    \hfill
    \subfigureCD{FMmicro}{fmm}{#1}


    \caption{#2}
    \label{fig:CD_#1}
  \end{figure*}
}

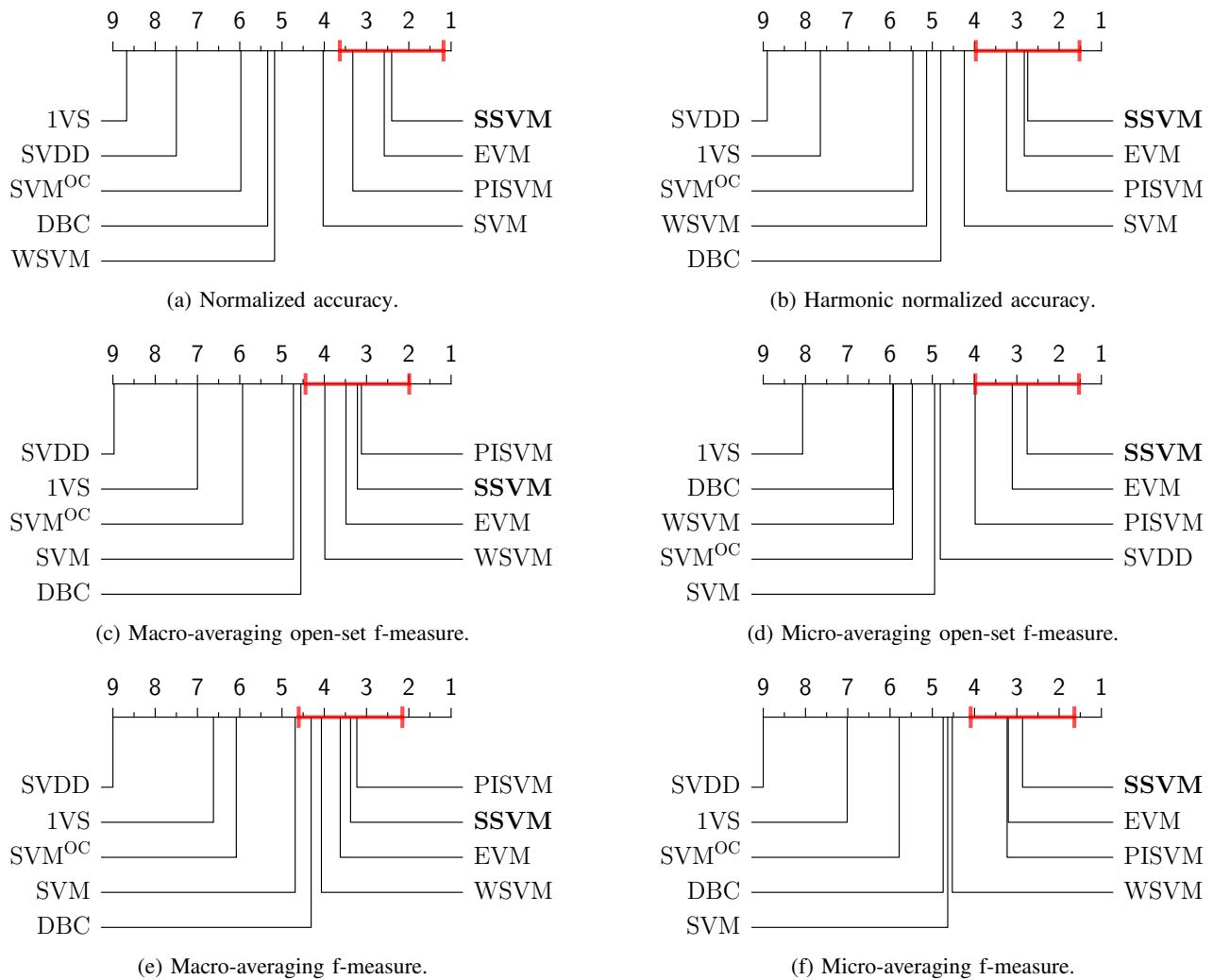
\figureCD{normal}{Critical difference diagram for comparison of \gls{ossvm} with baselines by employing Bonferroni-Dunn post-hoc test.  The bold-red bar along the horizontal numeric axis in each figure covers the methods with no statistical difference from \gls{ossvm}}

\subsection{Remarks}
\label{sec:remarks}

It is remarkable the frequency of negative bias terms in the binary classifiers that compose the \gls{mcfb} \gls{ova} \gls{svm}.
Most of the binary classifiers for the \gls{ova} approach already have the correct negative bias term, as shown in Table \ref{tab:correct-bias-term-frequency}.
To better understand the reason, we also obtained the frequency of binary classifiers with negative bias term using the \gls{ovo} approach, also shown in Table \ref{tab:correct-bias-term-frequency}.
In this case, only about half of the binary classifiers have negative bias term.

An informal explanation for this behavior is that in the \gls{ova} approach, we have more negative than positive samples---and more than one class in the negative set.
Then, it is more likely to have the negative samples ``surround'' the positive ones helping the \gls{svm} to create a bounded \gls{plos} for the positive class.
This intuition is confirmed by Figure \ref{fig:figPlottingkernel}.
For both class 1 and class 2, the \gls{svm} creates a bounded \gls{plos} (negative bias term) because class 3 (blue) is negative for those binary classifiers and it surrounds the positive class in both cases.
Considering class 3 as positive, we have no negative samples surround the positive class.
That is why, in this case, the \gls{plos} is unbounded (non-negative bias term).

The high frequency of negative bias term for the \gls{ova} explains why some authors in the literature have been reporting good accuracy for detection problems using \gls{svm}s with \gls{rbf} kernels.
For a detection problem, we have one class of interest and multiple others that we consider as negative for what we have access to train with.
As the number of negative samples is usually larger, it is more likely to have a classifier with bounded \gls{plos} for detection problems.

\tablePercentageNegativeBias{}

Despite most of the cases the \gls{svm} obtains the correct bias term for the \gls{ova} approach, the optimization problem presented in Section \ref{sec:ssvm} also optimizes the \gls{ru}.
That is, it optimizes for recognition in open-set scenario.

\section{Conclusions and Future Work}
\label{sec:conclusion}

\glsreset{smo}
\glsreset{wss}

In this work, we presented sufficient and necessary conditions for the \gls{svm} with \gls{rbf} kernel to have a finite \gls{ru}.
We then showed that by reformulating the \gls{rbf} \gls{svm} optimization policy to simultaneously optimize margin and ensure a negative bias term, the \gls{ru} is bounded and it provides a formal open-set recognition algorithm.

The proposed \gls{ossvm} method extends upon the traditional \gls{svm}'s optimization problem.
The objective function is changed in the primal problem, but the Lagrangian for the dual problem remains the same.
In the dual problem for the \gls{ossvm}, only a single constraint differs from the \gls{svm}'s dual problem.
Therefore, the same \gls{smo} algorithm \cite{Platt1998} can be used to ensure the new constraint is satisfied between iterations.
Also, the same \gls{wss} algorithm \cite{Fan2005} can be applied.

A limitation of the proposed method is that it can only be applied to specific kinds of kernel: the ones that tends to zero as the two given instances get far apart from each other.
Among the well-known kernels, only \gls{rbf} has this property, but others are available, e.g., \glsdesc{tst}, \glsdesc{rq}, and \glsdesc{imq} kernels \cite{Souza2010}.
Another limitation of this work is the lack of a proof that ensures the parameters selected during grid search phase can always generate a model, on the training phase, which has a negative bias term for bounding \gls{klos}.

As future work, one can investigate alternative forms for ensuring a bounded \gls{plos} for specific implementations of \gls{svm}s that do not rely on the bias term---as the ones in \citet{Vogt2002} and \citet{Kecman2005}.
As discussed in Section \ref{sec:ssvm-bounding-open-space}, the \gls{svm} without the bias term cannot ensure a bounded \gls{plos}, as it depends on the shape of the training data.
However, a simple solution can be obtained by training the \gls{svm} without bias term and establishing an artificial negative bias term in the decision function.
In this case, research can be done on how to obtain this artificial bias term.
Another alternative is to consider an optimization problem with a fixed negative bias term \cite{Platt1998}.

Another future work is to investigate properties of the \gls{mcfb} \gls{svm} with the \gls{ovo} approach.
In a binary \gls{svm}, at least one of the two classes will always have infinite \gls{plos}---if one is bounded, the other must include all the remaining space and so it must be infinite.
We observed experimentally that, according to the way the probability is calculated for each binary classifier \cite{Platt2000,Lin2007} and according to the way the probability estimates are combined in the \gls{mcfb} level \cite{Wu2004}, depending on the threshold established, a bounded \gls{klos} can occur but cannot be ensured.
Future work relies on investigating mechanisms to always ensure a bounded \gls{klos} for \gls{ovo} approach and, consequently, a limited \gls{ru}.
This is worth investigating because some works in the literature have presented better results with the \gls{ovo} than \gls{ova} approach for closed-set problems \cite{Galar2011}.
We then hypothesize that, for \gls{ovo} approach, this investigation should be accomplished in the probability estimation or/and probability combination.

Finally, the guarantees we present in this work are valid for the feature space of the feature description.
Ensuring a bounded mapping from the original feature space of the data to the space of feature description remains an open research problem associated to each description method.


%

\appendices
\section{Complete \texorpdfstring{\gls{ossvm}}{OSSVM} Formulation}
\label{appendix:sec:ssvm-formulation}

The optimization problem for the \gls{ossvm} classifier is defined as
\begin{align*}
  \min_{\mathbf{w},b,\xi}\ & \frac{1}{2}\|\mathbf{w}\|^{2} + C\sum_{i=1}^{m}\xi_{i} + \lambda b,\\
  \mbox{s.t.}\ & y_{i}\left(\mathbf{w}^{T}\mathbf{x}_{i} + b\right) - 1 + \xi_{i} \ge 0,\\
                           & \xi_{i} \ge 0.
\end{align*}

Using the Lagrangian method, we have the Lagrangian defined as
\begin{align}
  \label{appendix:eq:lagrangian-initial}
  \mathcal{L}(\mathbf{w}, b, \xi, \alpha, \mathbf{r}) &= \frac{1}{2}\|\mathbf{w}\|^{2} + C\sum_{i=1}^{m}\xi_{i} + \lambda b - \sum_{i=1}^{m}r_{i}\xi_{i} \nonumber \\
                                                      &- \sum_{i=1}^{m}\alpha_{i}\left[y_{i}\left(\mathbf{w}^{T}\mathbf{x}_{i} + b\right) - 1 + \xi_{i}\right],
\end{align}
in which $\alpha_{i} \in \mathbb{R}$ and $r_{i} \in \mathbb{R}$, $i = 1, \dots, m$, are the Lagrangian multipliers.

First we want to minimize with respect to $\mathbf{w}$, $b$, and $\xi_{i}$, then we must ensure
\begin{equation*}
  \nabla_{\mathbf{w}} = \frac{\partial}{\partial b}\mathcal{L} = \frac{\partial}{\partial\xi_{i}}\mathcal{L} = 0.
\end{equation*}

Consequently, we have
\begin{align}
  \label{appendix:eq:gradient-w}
  \mathbf{w} - \sum_{i=1}^{m}\alpha_{i}y_{i}\mathbf{x}_{i} = 0 \implies \mathbf{w} = \sum_{i=1}^{m}\alpha_{i}y_{i}\mathbf{x}_{i},\\
  \label{appendix:eq:derivative-b}
  \lambda - \sum_{i=1}^{m}\alpha_{i}y_{i} = 0 \implies \sum_{i=1}^{m}\alpha_{i}y_{i} = \lambda,\\
  \label{appendix:eq:derivative-xi}
  C - \alpha_{i} - r_{i} = 0 \implies r_{i} = C - \alpha_{i}.
\end{align}

As the Lagrangian multipliers $\alpha_{i}, r_{i}$ must be greater than 0, from Equation \eqref{appendix:eq:derivative-xi} we have the constraint $0 \le \alpha_{i} \le C$ as a consequence of the soft margin formulation in the dual problem.
This is the same constraint we have in the traditional formulation of the \gls{svm} classifier.

Using Equations \eqref{appendix:eq:gradient-w}--\eqref{appendix:eq:derivative-xi} to simplify the Lagrangian in Equation \eqref{appendix:eq:lagrangian-initial}, we have
\begin{align}
  \mathcal{L}(\mathbf{w}, b, \xi, \alpha) &= \frac{1}{2}\sum_{i=1}^{m}\sum_{j=1}^{m}\alpha_{i}\alpha_{i}y_{i}y_{j}\mathbf{x}_{i}^{T}\mathbf{x}_{j} + C\sum_{i=1}^{m}\xi_{i} \nonumber \\
                                          &+ b\sum_{i=1}^{m}\alpha_{i}y_{i} - C\sum_{i=1}^{m}\xi_{i} + \sum_{i=1}^{m}\alpha_{i}\xi_{i} \nonumber \\
                                          &- \sum_{i=1}^{m}\sum_{j=1}^{m}\alpha_{i}\alpha_{i}y_{i}y_{j}\mathbf{x}_{i}^{T}\mathbf{x}_{j} - b\sum_{i=1}^{m}\alpha_{i}y_{i} \nonumber \\
                                          &+ \sum_{i=1}^{m}\alpha_{i} - \sum_{i=1}^{m}\alpha_{i}\xi_{i}, \nonumber
\end{align}
which simplifies to
\begin{align}
  \mathcal{L}(\alpha) &= \sum_{i=1}^{m}\alpha_{i} - \frac{1}{2}\sum_{i=1}^{m}\sum_{j=1}^{m}\alpha_{i}\alpha_{i}y_{i}y_{j}\mathbf{x}_{i}^{T}\mathbf{x}_{j} \nonumber \\
  \label{appendix:eq:lagrangian}
                      &= \sum_{i=1}^{m}\alpha_{i} - \frac{1}{2}\|\mathbf{w}\|^{2}.
\end{align}

Equation~(\ref{appendix:eq:lagrangian}) shows the same Lagrangian of the traditional \gls{svm} optimization problem.  The optimization of the bias term $b$ relies on the constraint in Equation \eqref{appendix:eq:derivative-b}.

Therefore, the dual optimization problem is defined as
\begin{align*}
  \min_{\alpha}\ & \mathcal{W}(\alpha) = - \mathcal{L}(\alpha) = \frac{1}{2}\|\mathbf{w}\|^{2} - \sum_{i=1}^{m}\alpha_{i},\\
  \mbox{s.t.}\ & 0 < \alpha_{i} < C, \ \forall i,\\
                 & \sum_{i=1}^{m}\alpha_{i}y_{i} = \lambda.
\end{align*}

\section*{Acknowledgment}

This study was financed in part by the Coordenação de Aperfeiçoamento de Pessoal de Nível Superior---Brasil (CAPES)---Finance Code 001.
Part of the research were supported by CAPES through DeepEyes project and the scholarship provided to the first author.
The authors also thank the financial support of the Brazilian National Council for Scientific and Technological Development (CNPq),
the S{\~a}o Paulo Research Foundation (FAPESP) through D{\'e}j{\`a}Vu project, Grant \mbox{\#2017}/\mbox{12646-3},
and Microsoft Research.
This research is also based upon work funded in part by NSF \mbox{IIS-1320956}.
We also thank Brandon Richard Webster for the help on preparing \glsfirst{imagenet2010} features.
First author also thanks Bernardo Vecchia Stein for the collaboration on the initial analysis of \gls{svm} that led to the development of this work.
Finally, we thank all developers of the essential tools we have employed for accomplishing this work \cite[etc.]{Hunter2007,Perez2007,Pedregosa2011,Albanese2012,Demsar2013,Tange2018}.

\ifCLASSOPTIONcaptionsoff
  \newpage
\fi



\bibliographystyle{IEEEtranN}
\bibliography{mybib}

\begin{thebibliography}{86}
\providecommand{\natexlab}[1]{#1}
\providecommand{\url}[1]{#1}
\csname url@samestyle\endcsname
\providecommand{\newblock}{\relax}
\providecommand{\bibinfo}[2]{#2}
\providecommand{\BIBentrySTDinterwordspacing}{\spaceskip=0pt\relax}
\providecommand{\BIBentryALTinterwordstretchfactor}{4}
\providecommand{\BIBentryALTinterwordspacing}{\spaceskip=\fontdimen2\font plus
\BIBentryALTinterwordstretchfactor\fontdimen3\font minus
  \fontdimen4\font\relax}
\providecommand{\BIBforeignlanguage}[2]{{%
\expandafter\ifx\csname l@#1\endcsname\relax
\typeout{** WARNING: IEEEtranN.bst: No hyphenation pattern has been}%
\typeout{** loaded for the language `#1'. Using the pattern for}%
\typeout{** the default language instead.}%
\else
\language=\csname l@#1\endcsname
\fi
#2}}
\providecommand{\BIBdecl}{\relax}
\BIBdecl

\bibitem[Bishop(2006)]{Bishop2006}
C.~M. Bishop, \emph{Pattern Recognition and Machine Learning}, 1st~ed., ser.
  Information Science and Statistics.\hskip 1em plus 0.5em minus 0.4em\relax
  Springer-Verlag New York, 2006.

\bibitem[Breiman(2001)]{Breiman2001}
L.~Breiman, ``{R}andom {F}orests,'' \emph{Springer Machine Learning}, vol.~45,
  no.~1, pp. 5--32, Oct. 2001.

\bibitem[Cortes and Vapnik(1995)]{Cortes1995}
C.~Cortes and V.~Vapnik, ``Support-vector networks,'' \emph{Springer Machine
  Learning}, vol.~20, no.~3, pp. 273--297, Sep. 1995.

\bibitem[LeCun et~al.(2015)LeCun, Bengio, and Hinton]{LeCun2015}
Y.~LeCun, Y.~Bengio, and G.~Hinton, ``Deep learning,'' \emph{Nature}, vol. 521,
  no. 7553, pp. 436--444, May 2015.

\bibitem[Costa et~al.(2014)Costa, Silva, Eckmann, Scheirer, and
  Rocha]{Costa2014}
F.~d.~O. Costa, E.~Silva, M.~Eckmann, W.~J. Scheirer, and A.~Rocha, ``Open set
  source camera attribution and device linking,'' \emph{Elsevier Pattern
  Recognition Letters}, vol.~39, pp. 92--101, Apr. 2014.

\bibitem[Ferreira et~al.(2017)Ferreira, Bondi, Baroffio, Bestagini, Huang, dos
  Santos, Tubaro, and Rocha]{Ferreira2017}
A.~Ferreira, L.~Bondi, L.~Baroffio, P.~Bestagini, J.~Huang, J.~A. dos Santos,
  S.~Tubaro, and A.~Rocha, ``Data-driven feature characterization techniques
  for laser printer attribution,'' \emph{IEEE Transactions on Information
  Forensics and Security}, vol.~12, no.~8, pp. 1860--1873, Aug. 2017.

\bibitem[Dubuisson and Masson(1993)]{Dubuisson1993}
B.~Dubuisson and M.~Masson, ``A statistical decision rule with incomplete
  knowledge about classes,'' \emph{Elsevier Pattern Recognition}, vol.~26,
  no.~1, pp. 155--165, Jan. 1993.

\bibitem[Muzzolini et~al.(1998)Muzzolini, Yang, and Pierson]{Muzzolini1998}
R.~Muzzolini, Y.-H. Yang, and R.~Pierson, ``Classifier design with incomplete
  knowledge,'' \emph{Elsevier Pattern Recognition}, vol.~31, no.~4, pp.
  345--369, Apr. 1998.

\bibitem[Moeini et~al.(2017)Moeini, Faez, Moeini, and Safai]{Moeini2017}
A.~Moeini, K.~Faez, H.~Moeini, and A.~M. Safai, ``Open-set face recognition
  across look-alike faces in real-world scenarios,'' \emph{Elsevier Image and
  Vision Computing}, vol.~57, pp. 1--14, Jan. 2017.

\bibitem[Mendes~Júnior et~al.(2017)Mendes~Júnior, de~Souza, Werneck, Stein,
  Pazinato, de~Almeida, Penatti, Torres, and Rocha]{MendesJunior2017}
P.~R. Mendes~Júnior, R.~M. de~Souza, R.~d.~O. Werneck, B.~V. Stein, D.~V.
  Pazinato, W.~R. de~Almeida, O.~A.~B. Penatti, R.~d.~S. Torres, and A.~d.~R.
  Rocha, ``Nearest neighbors distance ratio open-set classifier,''
  \emph{Springer Machine Learning}, vol. 106, no.~3, pp. 359--386, Mar. 2017.

\bibitem[Liu et~al.(2017)Liu, Wen, Yu, Li, Raj, and Song]{Liu2017}
W.~Liu, Y.~Wen, Z.~Yu, M.~Li, B.~Raj, and L.~Song, ``{S}phere{F}ace: Deep
  hypersphere embedding for face recognition,'' in \emph{IEEE International
  Conference on Computer Vision and Pattern Recognition}, Honolulu, HI, USA,
  Jul. 2017, pp. 212--220.

\bibitem[Wen et~al.(2019)Wen, Zhang, Li, and Qiao]{Wen2019}
Y.~Wen, K.~Zhang, Z.~Li, and Y.~Qiao, ``A comprehensive study on center loss
  for deep face recognition,'' \emph{Springer International Journal of Computer
  Vision}, vol. 127, no. 6--7, pp. 668--683, Jun. 2019.

\bibitem[Tax and Duin(2004)]{Tax2004}
D.~M.~J. Tax and R.~P.~W. Duin, ``Support vector data description,''
  \emph{Springer Machine Learning}, vol.~54, no.~1, pp. 45--66, Jan. 2004.

\bibitem[Chang et~al.(2013)Chang, Lee, and Lin]{Chang2013}
W.-C. Chang, C.-P. Lee, and C.-J. Lin, ``A revisit to {S}upport {V}ector {D}ata
  {D}escription,'' National Taiwan University of Science and Technology,
  Taipei, Taiwan, Tech. Rep., 2013.

\bibitem[Schölkopf et~al.(2001)Schölkopf, Platt, Shawe-Taylor, Smola, and
  Williamson]{Scholkopf2001a}
B.~Schölkopf, J.~C. Platt, J.~Shawe-Taylor, A.~J. Smola, and R.~C. Williamson,
  ``Estimating the support of a high-dimensional distribution,'' \emph{Neural
  Computation}, vol.~13, no.~7, pp. 1443--1471, Jul. 2001.

\bibitem[Cortes et~al.(2016)Cortes, DeSalvo, and Mohri]{Cortes2016}
C.~Cortes, G.~DeSalvo, and M.~Mohri, ``Learning with rejection,'' in
  \emph{International Conference on Algorithmic Learning Theory}, ser. Lecture
  Notes in Computer Science, R.~Ortner, H.~U. Simon, and S.~Zilles, Eds., vol.
  9925.\hskip 1em plus 0.5em minus 0.4em\relax Bari, Italy: Springer
  International Publishing, Oct. 2016, pp. 67--82.

\bibitem[Rocha and Goldenstein(2014)]{Rocha2014}
A.~Rocha and S.~Goldenstein, ``Multiclass from binary: Expanding one-vs-all,
  one-vs-one and {ECOC}-based approaches,'' \emph{IEEE Transactions on Neural
  Networks and Learning Systems}, vol.~25, no.~2, pp. 289--302, Feb. 2014.

\bibitem[Pritsos and Stamatatos(2013)]{Pritsos2013}
D.~A. Pritsos and E.~Stamatatos, ``Open-set classification for automated genre
  identification,'' in \emph{European Conference on Information Retrieval},
  ser. Lecture Notes in Computer Science, P.~Serdyukov, P.~Braslavski, S.~O.
  Kuznetsov, J.~Kamps, and S.~Rüger, Eds., vol. 7814.\hskip 1em plus 0.5em
  minus 0.4em\relax Moscow, Russia: Springer, Berlin, Heidelberg, Mar. 2013,
  pp. 207--217.

\bibitem[Scheirer et~al.(2014)Scheirer, Jain, and Boult]{Scheirer2014}
W.~J. Scheirer, L.~P. Jain, and T.~E. Boult, ``Probability models for open set
  recognition,'' \emph{IEEE Transactions on Pattern Analysis and Machine
  Intelligence}, vol.~36, no.~11, pp. 2317--2324, Nov. 2014.

\bibitem[Jain et~al.(2014)Jain, Scheirer, and Boult]{Jain2014}
L.~P. Jain, W.~J. Scheirer, and T.~E. Boult, ``Multi-class open set recognition
  using probability of inclusion,'' in \emph{European Conference on Computer
  Vision}, ser. Lecture Notes in Computer Science, D.~Fleet, T.~Pajdla,
  B.~Schiele, and T.~Tuytelaars, Eds., vol. 8691, part III.\hskip 1em plus
  0.5em minus 0.4em\relax Zurich, Switzerland: Springer, Cham, Sep. 2014, pp.
  393--409.

\bibitem[Scheirer et~al.(2013)Scheirer, Rocha, Sapkota, and
  Boult]{Scheirer2013}
W.~J. Scheirer, A.~d.~R. Rocha, A.~Sapkota, and T.~E. Boult, ``Towards open set
  recognition,'' \emph{IEEE Transactions on Pattern Analysis and Machine
  Intelligence}, vol.~35, no.~7, pp. 1757--1772, Jul. 2013.

\bibitem[Heflin et~al.(2012)Heflin, Scheirer, and Boult]{Heflin2012}
B.~Heflin, W.~J. Scheirer, and T.~E. Boult, ``Detecting and classifying scars,
  marks, and tattoos found in the wild,'' in \emph{IEEE International
  Conference on Biometrics: Theory, Applications and Systems}, Arlington, VA,
  USA, Sep. 2012, pp. 31--38.

\bibitem[Mendes~Júnior et~al.(2019)Mendes~Júnior, Bondi, Bestagini, Tubaro,
  and Rocha]{MendesJunior2019}
P.~R. Mendes~Júnior, L.~Bondi, P.~Bestagini, S.~Tubaro, and A.~Rocha, ``An
  in-depth study on open-set camera model identification,'' \emph{IEEE Access},
  vol.~7, pp. 180\,713--180\,726, Jun. 2019.

\bibitem[Costa et~al.(2012)Costa, Eckmann, Scheirer, and Rocha]{Costa2012}
F.~d.~O. Costa, M.~Eckmann, W.~J. Scheirer, and A.~Rocha, ``Open set source
  camera attribution,'' in \emph{Conference on Graphics, Patterns, and
  Images}.\hskip 1em plus 0.5em minus 0.4em\relax Ouro Preto, MG, Brazil: IEEE
  Press, Aug. 2012, pp. 71--78.

\bibitem[Coles(2001)]{Coles2001}
S.~Coles, \emph{An Introduction to Statistical Modeling of Extreme Values},
  1st~ed., ser. Springer Series in Statistics.\hskip 1em plus 0.5em minus
  0.4em\relax Springer, London, 2001.

\bibitem[Rudd et~al.(2018)Rudd, Jain, Scheirer, and Boult]{Rudd2018}
E.~M. Rudd, L.~P. Jain, W.~J. Scheirer, and T.~E. Boult, ``The {E}xtreme
  {V}alue {M}achine,'' \emph{IEEE Transactions on Pattern Analysis and Machine
  Intelligence}, vol.~40, no.~3, pp. 762--768, Mar. 2018.

\bibitem[Scheirer(2017)]{Scheirer2017}
W.~J. Scheirer, \emph{Extreme Value Theory-Based Methods for Visual
  Recognition}, 1st~ed., ser. Synthesis Lectures on Computer Vision.\hskip 1em
  plus 0.5em minus 0.4em\relax Morgan \& Claypool Publishers, Feb. 2017.

\bibitem[Vareto et~al.(2017)Vareto, Silva, Costa, and Schwartz]{Vareto2017}
R.~Vareto, S.~Silva, F.~Costa, and W.~R. Schwartz, ``Towards open-set face
  recognition using hashing functions,'' in \emph{IEEE International Joint
  Conference on Biometrics}, Denver, CO, USA, Oct. 2017, pp. 634--641.

\bibitem[Busto and Gall(2017)]{Busto2017}
P.~P. Busto and J.~Gall, ``Open set domain adaptation,'' in \emph{IEEE
  International Conference on Computer Vision}, Venice, Italy, Oct. 2017, pp.
  754--763.

\bibitem[Saito et~al.(2018)Saito, Yamamoto, Ushiku, and Harada]{Saito2018}
K.~Saito, S.~Yamamoto, Y.~Ushiku, and T.~Harada, ``Open set domain adaptation
  by backpropagation,'' in \emph{European Conference on Computer Vision}, ser.
  Lecture Notes in Computer Science, V.~Ferrari, M.~Hebert, C.~Sminchisescu,
  and Y.~Weiss, Eds., vol. 11205.\hskip 1em plus 0.5em minus 0.4em\relax
  Munich, Germany: Springer, Cham, Sep. 2018, pp. 156--171.

\bibitem[Fu et~al.(2019)Fu, Wu, Zhang, and Yan]{Fu2019}
J.~Fu, X.~Wu, S.~Zhang, and J.~Yan, ``Improved open set domain adaptation with
  backpropagation,'' in \emph{IEEE International Conference on Image
  Processing}, Taipei, Taiwan, Sep. 2019, pp. 2506--2510.

\bibitem[Liu et~al.(2019)Liu, Cao, Long, Wang, and Yang]{Liu2019b}
H.~Liu, Z.~Cao, M.~Long, J.~Wang, and Q.~Yang, ``Separate to adapt: Open set
  domain adaptation via progressive separation,'' in \emph{IEEE International
  Conference on Computer Vision and Pattern Recognition}.\hskip 1em plus 0.5em
  minus 0.4em\relax Long Beach, CA, USA: {IEEE}, Jun. 2019, pp. 2922--2931.

\bibitem[Neira et~al.(2018)Neira, Mendes~Júnior, Rocha, and Torres]{Neira2018}
M.~A.~C. Neira, P.~R. Mendes~Júnior, A.~Rocha, and R.~d.~S. Torres,
  ``Data-fusion techniques for open-set recognition problems,'' \emph{IEEE
  Access}, vol.~6, pp. 21\,242--21\,265, Apr. 2018.

\bibitem[Dhamija et~al.(2020)Dhamija, Günther, Ventura, and
  Boult]{Dhamija2020}
A.~Dhamija, M.~Günther, J.~Ventura, and T.~Boult, ``The overlooked elephant of
  object detection: Open set,'' in \emph{Winter Conference on Applications of
  Computer Vision}.\hskip 1em plus 0.5em minus 0.4em\relax Aspen, CO, USA: IEEE
  Press, Mar. 2020, pp. 1021--1030.

\bibitem[Dang et~al.(2019)Dang, Cao, Cui, Pi, and Liu]{Dang2019}
S.~Dang, Z.~Cao, Z.~Cui, Y.~Pi, and N.~Liu, ``Open set incremental learning for
  automatic target recognition,'' \emph{IEEE Transactions on Geoscience and
  Remote Sensing}, vol.~57, no.~7, pp. 4445--4456, Jul. 2019.

\bibitem[Moraes et~al.(2016)Moraes, Wainer, and Rocha]{Moraes2016}
D.~Moraes, J.~Wainer, and A.~Rocha, ``Low false positive learning with
  {S}upport {V}ector {M}achines,'' \emph{Elsevier Journal of Visual
  Communication and Image Representation}, vol.~38, pp. 340--350, Jul. 2016.

\bibitem[Tax and Duin(2008)]{Tax2008}
D.~M.~J. Tax and R.~P.~W. Duin, ``Growing a multi-class classifier with a
  reject option,'' \emph{Elsevier Pattern Recognition Letters}, vol.~29,
  no.~10, pp. 1565--1570, Jul. 2008.

\bibitem[Bartlett and Wegkamp(2008)]{Bartlett2008}
P.~L. Bartlett and M.~H. Wegkamp, ``Classification with a reject option using a
  hinge loss,'' \emph{Journal of Machine Learning Research}, vol.~9, pp.
  1823--1840, Aug. 2008.

\bibitem[Saki et~al.(2019)Saki, Guo, Hung, Kim, Deshpande, Moon, Koh, and
  Visser]{Saki2019}
F.~Saki, Y.~Guo, C.-Y. Hung, L.-h. Kim, M.~Deshpande, S.~Moon, E.~Koh, and
  E.~Visser, ``Open-set evolving acoustic scene classification system,'' in
  \emph{Detection and Classification of Acoustic Scenes and Events Workshop},
  New York, NY, USA, Oct. 2019, pp. 219--223.

\bibitem[Tornai and Scheirer(2019)]{Tornai2019}
K.~Tornai and W.~J. Scheirer, ``Gesture-based user identity verification as an
  open set problem for smartphones,'' in \emph{IAPR International Conference on
  Biometrics}, Crete, Greece, Jun. 2019, pp. 1--8.

\bibitem[Yang et~al.(2019)Yang, Hou, Lang, Guan, Huang, and Xu]{Yang2019a}
Y.~Yang, C.~Hou, Y.~Lang, D.~Guan, D.~Huang, and J.~Xu, ``Open-set human
  activity recognition based on micro-{D}oppler signatures,'' \emph{Elsevier
  Pattern Recognition}, vol.~85, pp. 60--69, Jan. 2019.

\bibitem[Geng et~al.(2020)Geng, Huang, and Chen]{Geng2020}
C.~Geng, S.-J. Huang, and S.~Chen, ``Recent advances in open set recognition: A
  survey,'' \emph{IEEE Transactions on Pattern Analysis and Machine
  Intelligence}, pp. 1--19, Mar. 2020, early Access.

\bibitem[Boser et~al.(1992)Boser, Guyon, and Vapnik]{Boser1992}
B.~E. Boser, I.~M. Guyon, and V.~N. Vapnik, ``A training algorithm for optimal
  margin classifiers,'' in \emph{ACM Annual Workshop on Computational Learning
  Theory}, Pittsburgh, PA, USA, Jul. 1992, pp. 144--152.

\bibitem[Mercer(1909)]{Mercer1909}
J.~Mercer, ``Functions of positive and negative type, and their connection the
  theory of integral equations,'' \emph{Philosophical Transactions of the Royal
  Society A}, vol. 209, no. 441--458, pp. 415--446, Jan. 1909.

\bibitem[Schölkopf and Smola(2001)]{Scholkopf2001}
B.~Schölkopf and A.~J. Smola, \emph{Learning with Kernels}, 1st~ed., ser.
  Adaptive Computation and Machine Learning series.\hskip 1em plus 0.5em minus
  0.4em\relax MIT Press, Dec. 2001.

\bibitem[Chang and Lin(2011)]{Chang2011}
C.-C. Chang and C.-J. Lin, ``{LIBSVM}: A library for {S}upport {V}ector
  {M}achines,'' \emph{ACM Transactions on Intelligent Systems and Technology},
  vol.~2, no.~3, pp. 27:1--27:27, Apr. 2011.

\bibitem[Buhmann(2003)]{Buhmann2003}
M.~D. Buhmann, \emph{Radial Basis Functions: Theory and Implementations},
  1st~ed., ser. Cambridge Monographs on Applied and Computational
  Mathematics.\hskip 1em plus 0.5em minus 0.4em\relax Cambridge University
  Press, Jul. 2003.

\bibitem[Souza(2010)]{Souza2010}
\BIBentryALTinterwordspacing
C.~R. Souza, ``Kernel functions for machine learning applications,'' Mar. 2010,
  accessed: November 14, 2018. [Online]. Available:
  \url{http://crsouza.com/2010/03/17/kernel-functions-for-machine-learning-applications}
\BIBentrySTDinterwordspacing

\bibitem[Vogt(2002)]{Vogt2002}
M.~Vogt, ``{SMO} algorithms for {S}upport {V}ector {M}achines without bias
  term,'' Institute of Automatic Control, Technische Universität Darmstadt,
  Darmstadt, Germany, Tech. Rep., Jul. 2002.

\bibitem[Kecman et~al.(2005)Kecman, Huang, and Vogt]{Kecman2005}
V.~Kecman, T.~M. Huang, and M.~Vogt, ``Iterative single data algorithm for
  training kernel machines from huge data sets: Theory and performance,'' in
  \emph{{S}upport {V}ector {M}achines: Theory and Applications}, ser. Studies
  in Fuzziness and Soft Computing, L.~Wang, Ed.\hskip 1em plus 0.5em minus
  0.4em\relax Springer, Berlin, Heidelberg, Apr. 2005, vol. 177, pp. 255--274.

\bibitem[Kuncheva and Hadjitodorov(2004)]{Kuncheva2004}
L.~I. Kuncheva and S.~T. Hadjitodorov, ``Using diversity in cluster
  ensembles,'' in \emph{IEEE International Conference on Systems, Man, and
  Cybernetics}, vol.~2, The Hague, Netherlands, Oct. 2004, pp. 1214--1219.

\bibitem[Platt(1998)]{Platt1998}
J.~C. Platt, ``Fast training of {S}upport {V}ector {M}achines using
  {S}equential {M}inimal {O}ptimization,'' in \emph{Advances in Kernel Methods:
  Support Vector Learning}, 1st~ed., B.~Schölkopf, C.~J.~C. Burges, and A.~J.
  Smola, Eds.\hskip 1em plus 0.5em minus 0.4em\relax MIT Press, Dec. 1998,
  ch.~12, pp. 185--208.

\bibitem[Fan et~al.(2005)Fan, Chen, and Lin]{Fan2005}
R.-E. Fan, P.-H. Chen, and C.-J. Lin, ``{W}orking {S}et {S}election using
  second order information for training {S}upport {V}ector {M}achines,''
  \emph{Journal of Machine Learning Research}, vol.~6, pp. 1889--1918, Dec.
  2005.

\bibitem[Sokolova and Lapalme(2009)]{Sokolova2009}
M.~Sokolova and G.~Lapalme, ``A systematic analysis of performance measures for
  classification tasks,'' \emph{Elsevier Information Processing \& Management},
  vol.~45, no.~4, pp. 427--437, Jul. 2009.

\bibitem[Mitchell(2004)]{Mitchell2004}
D.~W. Mitchell, ``More on spreads and non-arithmetic means,'' \emph{The
  Mathematical Gazette}, vol.~88, no. 511, pp. 142--144, Mar. 2004.

\bibitem[Lazebnik et~al.(2006)Lazebnik, Schmid, and Ponce]{Lazebnik2006}
S.~Lazebnik, C.~Schmid, and J.~Ponce, ``Beyond bags of features: Spatial
  pyramid matching for recognizing natural scene categories,'' in \emph{IEEE
  International Conference on Computer Vision and Pattern Recognition}, vol.~2,
  New York, NY, USA, Jun. 2006, pp. 2169--2178.

\bibitem[van Gemert et~al.(2010)van Gemert, Veenman, Smeulders, and
  Geusebroek]{vanGemert2010}
J.~C. van Gemert, C.~J. Veenman, A.~W.~M. Smeulders, and J.-M. Geusebroek,
  ``Visual word ambiguity,'' \emph{IEEE Transactions on Pattern Analysis and
  Machine Intelligence}, vol.~32, no.~7, pp. 1271--1283, Jul. 2010.

\bibitem[Boureau et~al.(2010)Boureau, Bach, LeCun, and Ponce]{Boureau2010}
Y.-L. Boureau, F.~Bach, Y.~LeCun, and J.~Ponce, ``Learning mid-level features
  for recognition,'' in \emph{IEEE International Conference on Computer Vision
  and Pattern Recognition}, San Francisco, CA, USA, Jun. 2010, pp. 2559--2566.

\bibitem[Lowe(2004)]{Lowe2004}
D.~Lowe, ``Distinctive image features from scale-invariant keypoints,''
  \emph{Springer International Journal of Computer Vision}, vol.~60, no.~2, pp.
  91--110, Nov. 2004.

\bibitem[Frey and Slate(1991)]{Frey1991}
P.~W. Frey and D.~J. Slate, ``Letter recognition using {H}olland-style adaptive
  classifiers,'' \emph{Springer Machine Learning}, vol.~6, no.~2, pp. 161--182,
  Mar. 1991.

\bibitem[Michie et~al.(1994)Michie, Spiegelhalter, and Taylor]{Michie1994}
D.~Michie, D.~J. Spiegelhalter, and C.~C. Taylor, \emph{Machine Learning,
  Neural and Statistical Classification}, ser. Ellis Horwood Series in
  Artificial Intelligence.\hskip 1em plus 0.5em minus 0.4em\relax Upper Saddle
  River, NJ, USA: Prentice Hall, Jul. 1994.

\bibitem[Stolfo et~al.(2000)Stolfo, Fan, Lee, Prodromidis, and
  Chan]{Stolfo2000}
S.~J. Stolfo, W.~Fan, W.~Lee, A.~Prodromidis, and P.~K. Chan, ``Cost-based
  modeling for fraud and intrusion detection: Results from the {JAM} project,''
  in \emph{DARPA Information Survivability Conference and Exposition},
  vol.~2.\hskip 1em plus 0.5em minus 0.4em\relax Hilton Head, SC, USA: IEEE
  Press, Jan. 2000, pp. 130--144.

\bibitem[Kadous(2002)]{Kadous2002}
M.~W. Kadous, ``Temporal classification: Extending the classification paradigm
  to multivariate time series,'' Ph.D. dissertation, The University of New
  South Wales, New South Wales, Australia, Oct. 2002.

\bibitem[Griffin et~al.(2007)Griffin, Holub, and Perona]{Griffin2007}
G.~Griffin, A.~Holub, and P.~Perona, ``Caltech-256 object category dataset,''
  California Institute of Technology, Tech. Rep., May 2007.

\bibitem[Geusebroek et~al.(2005)Geusebroek, Burghouts, and
  Smeulders]{Geusebroek2005}
J.-M. Geusebroek, G.~J. Burghouts, and A.~W.~M. Smeulders, ``The {A}msterdam
  library of object images,'' \emph{Springer International Journal of Computer
  Vision}, vol.~61, no.~1, pp. 103--112, Jan. 2005.

\bibitem[Stehling et~al.(2002)Stehling, Nascimento, and Falcão]{Stehling2002}
R.~O. Stehling, M.~A. Nascimento, and A.~X. Falcão, ``A compact and efficient
  image retrieval approach based on border/interior pixel classification,'' in
  \emph{ACM International Conference on Information and Knowledge Management},
  McLean, VA, USA, Nov. 2002, pp. 102--109.

\bibitem[Russakovsky et~al.(2015)Russakovsky, Deng, Su, Krause, Satheesh, Ma,
  Huang, Karpathy, Khosla, Bernstein, Berg, and Fei-Fei]{Russakovsky2015}
O.~Russakovsky, J.~Deng, H.~Su, J.~Krause, S.~Satheesh, S.~Ma, Z.~Huang,
  A.~Karpathy, A.~Khosla, M.~Bernstein, A.~C. Berg, and L.~Fei-Fei,
  ``{I}mage{N}et large scale visual recognition challenge,'' \emph{Springer
  International Journal of Computer Vision}, vol. 115, no.~3, pp. 211--252,
  Dec. 2015.

\bibitem[Bendale and Boult(2016)]{Bendale2016}
A.~Bendale and T.~E. Boult, ``Towards open set deep networks,'' in \emph{IEEE
  International Conference on Computer Vision and Pattern Recognition}, Las
  Vegas, NV, USA, Jun. 2016, pp. 1563--1572.

\bibitem[Szegedy et~al.(2015)Szegedy, Liu, Jia, Sermanet, Reed, Anguelov,
  Erhan, Vanhoucke, and Rabinovich]{Szegedy2015}
C.~Szegedy, W.~Liu, Y.~Jia, P.~Sermanet, S.~Reed, D.~Anguelov, D.~Erhan,
  V.~Vanhoucke, and A.~Rabinovich, ``Going deeper with convolutions,'' in
  \emph{IEEE International Conference on Computer Vision and Pattern
  Recognition}, Boston, MA, USA, Jun. 2015, pp. 1--9.

\bibitem[Tipping and Bishop(1999)]{Tipping1999}
M.~E. Tipping and C.~M. Bishop, ``Probabilistic principal component analysis,''
  \emph{Journal of the Royal Statistical Society: Series B (Statistical
  Methodology)}, vol.~61, no.~3, pp. 611--622, 1999.

\bibitem[Krizhevsky and Hinton(2009)]{Krizhevsky2009}
A.~Krizhevsky and G.~Hinton, ``Learning multiple layers of features from tiny
  images,'' Master's thesis, University of Toronto, Apr. 2009.

\bibitem[LeCun et~al.(1998)LeCun, Bottou, Bengio, and Haffner]{LeCun1998}
Y.~LeCun, L.~Bottou, Y.~Bengio, and P.~Haffner, ``Gradient-based learning
  applied to document recognition,'' \emph{Proceedings of the IEEE}, vol.~86,
  no.~11, pp. 2278--2324, Nov. 1998.

\bibitem[Tensorflow.org(2018{\natexlab{a}})]{TensorflowCIFAR10}
\BIBentryALTinterwordspacing
Tensorflow.org, ``Advanced {C}onvolutional {N}eural {N}etworks,'' Oct. 2018,
  accessed: November 14, 2018. [Online]. Available:
  \url{https://www.tensorflow.org/tutorials/images/deep_cnn}
\BIBentrySTDinterwordspacing

\bibitem[Tensorflow.org(2018{\natexlab{b}})]{TensorflowDeepMNIST}
\BIBentryALTinterwordspacing
------, ``Deep {MNIST} for experts,'' May 2018, backup version. Accessed:
  November 14, 2018. [Online]. Available:
  \url{https://apimirror.com/tensorflow~guide/get_started/mnist/pros}
\BIBentrySTDinterwordspacing

\bibitem[Olson et~al.(2017)Olson, La~Cava, Orzechowski, Urbanowicz, and
  Moore]{Olson2017}
R.~S. Olson, W.~La~Cava, P.~Orzechowski, R.~J. Urbanowicz, and J.~H. Moore,
  ``{PMLB}: A large benchmark suite for machine learning evaluation and
  comparison,'' \emph{BioData Mining}, vol.~10, no.~36, pp. 1--13, Dec. 2017.

\bibitem[Dem{\v{s}}ar(2006)]{Demsar2006}
J.~Dem{\v{s}}ar, ``Statistical comparisons of classifiers over multiple data
  sets,'' \emph{Journal of Machine Learning Research}, vol.~7, pp. 1--30, Jan.
  2006.

\bibitem[Platt(2000)]{Platt2000}
J.~C. Platt, ``Probabilities for {SV} {M}achines,'' in \emph{Advances in
  Large-Margin Classifiers}, ser. Neural Information Processing series, A.~J.
  Smola, P.~Bartlett, B.~Schölkopf, and D.~Schuurmans, Eds.\hskip 1em plus
  0.5em minus 0.4em\relax MIT Press, Sep. 2000, ch.~5, pp. 61--74.

\bibitem[Lin et~al.(2007)Lin, Lin, and Weng]{Lin2007}
H.-T. Lin, C.-J. Lin, and R.~C. Weng, ``A note on {P}latt's probabilistic
  outputs for {S}upport {V}ector {M}achines,'' \emph{Springer Machine
  Learning}, vol.~68, no.~3, pp. 267--276, Oct. 2007.

\bibitem[Wu et~al.(2004)Wu, Lin, and Weng]{Wu2004}
T.-F. Wu, C.-J. Lin, and R.~C. Weng, ``Probability estimates for multi-class
  classification by pairwise coupling,'' \emph{Journal of Machine Learning
  Research}, vol.~5, pp. 975--1005, Aug. 2004.

\bibitem[Galar et~al.(2011)Galar, Fernández, Barrenechea, Bustince, and
  Herrera]{Galar2011}
M.~Galar, A.~Fernández, E.~Barrenechea, H.~Bustince, and F.~Herrera, ``An
  overview of ensemble methods for binary classifiers in multi-class problems:
  Experimental study on one-vs-one and one-vs-all schemes,'' \emph{Elsevier
  Pattern Recognition}, vol.~44, no.~8, pp. 1761--1776, Aug. 2011.

\bibitem[Hunter(2007)]{Hunter2007}
J.~D. Hunter, ``{M}atplotlib: A {2D} graphics environment,'' \emph{Computing in
  Science \& Engineering}, vol.~9, no.~3, pp. 90--95, Jun. 2007.

\bibitem[Perez and Granger(2007)]{Perez2007}
F.~Perez and B.~E. Granger, ``{IP}ython: A system for interactive scientific
  computing,'' \emph{Computing in Science \& Engineering}, vol.~9, no.~3, pp.
  21--29, Jun. 2007.

\bibitem[Pedregosa et~al.(2011)Pedregosa, Varoquaux, Gramfort, Michel, Thirion,
  Grisel, Blondel, Prettenhofer, Weiss, Dubourg, Vanderplas, Passos,
  Cournapeau, Brucher, Perrot, and Duchesnay]{Pedregosa2011}
F.~Pedregosa, G.~Varoquaux, A.~Gramfort, V.~Michel, B.~Thirion, O.~Grisel,
  M.~Blondel, P.~Prettenhofer, R.~Weiss, V.~Dubourg, J.~Vanderplas, A.~Passos,
  D.~Cournapeau, M.~Brucher, M.~Perrot, and {\'{E}}.~Duchesnay, ``Scikit-learn:
  Machine learning in {P}ython,'' \emph{Journal of Machine Learning Research},
  vol.~12, pp. 2825--2830, Oct. 2011.

\bibitem[Albanese et~al.(2012)Albanese, Visintainer, Merler, Riccadonna,
  Jurman, and Furlanello]{Albanese2012}
D.~Albanese, R.~Visintainer, S.~Merler, S.~Riccadonna, G.~Jurman, and
  C.~Furlanello, ``{mlpy}: {M}achine {L}earning {P}ython,'' \emph{CoRR}, vol.
  abs/1202.6548, pp. 1--4, Mar. 2012.

\bibitem[Dem{\v{s}}ar et~al.(2013)Dem{\v{s}}ar, Curk, Erjavec, Gorup,
  Ho{\v{c}}evar, Milutinovi{\v{c}}, Mo{\v{z}}ina, Polajnar, Toplak,
  Stari{\v{c}}, {\v{S}}tajdohar, Umek, {\v{Z}}agar, {\v{Z}}bontar,
  {\v{Z}}itnik, and Zupan]{Demsar2013}
J.~Dem{\v{s}}ar, T.~Curk, A.~Erjavec, {\v{C}}.~Gorup, T.~Ho{\v{c}}evar,
  M.~Milutinovi{\v{c}}, M.~Mo{\v{z}}ina, M.~Polajnar, M.~Toplak,
  A.~Stari{\v{c}}, M.~{\v{S}}tajdohar, L.~Umek, L.~{\v{Z}}agar,
  J.~{\v{Z}}bontar, M.~{\v{Z}}itnik, and B.~Zupan, ``{O}range: Data mining
  toolbox in {P}ython,'' \emph{Journal of Machine Learning Research}, vol.~14,
  pp. 2349--2353, Aug. 2013.

\bibitem[Tange(2018)]{Tange2018}
O.~Tange, ``{GNU} parallel 2018,'' GNU's Not Unix, Tech. Rep., Apr. 2018.

\end{thebibliography}
%



%

\begin{IEEEbiography}[{\includegraphics[width=1in,height=1.25in,clip,keepaspectratio]{\changedir{bio_photos/}MendesJunior}}]{Pedro Ribeiro Mendes J\'{u}nior}
is a postdoctoral researcher at Institute of Computing at the University of Campinas.  He received his PhD (2018) and MSc (2014) in Computer Science by the same university.  He has passed part of his PhD in an internship at University of Colorado, Colorado Springs, with the supervision of Prof.\@ Terrance Boult, period in which the OSSVM method was most formalized.  He stayed for a short post-doctoral internship at Politecnico di Milano working with Prof.\@ Stefano Tubaro and Prof.\@ Paolo Bestagini, period in which OSSVM was applied with success for other applications.  He obtained his Bachelor’s in Computer Science at the Federal University of Ouro Preto (Universidade Federal de Ouro Preto), period in which he has worked mainly on image processing and computer vision.  His current research topics are open-set recognition, computer vision, neural networks, time-series forecasting, time-series anomaly detection, etc. and on living consciously.
\end{IEEEbiography}

\begin{IEEEbiography}[{\includegraphics[width=1in,height=1.25in,clip,keepaspectratio]{\changedir{bio_photos/}boult-terry}}]{Terrance E. Boult}
is a Distinguished Professor and the El Pomar Endowed Professor of Innovation and Security at U.\@ Colorado Colorado Springs, as well as being an IEEE fellow and a serial entrepreneur and internationally acknowledged researcher in machine learning, computer vision, biometrics, and cybersecurity with 15 patents issued, 400+ papers.  He received his BS in Applied Math (1983), MS in CS (1984), and Ph.D.\@ in Computer CS (1986) from Columbia University and then spent six years as an Assistant Prof and two years as an Assoc.\@ Prof in Columbia's CS Department.  He moved from Columbia to Lehigh (1994-2003), where he was an endowed professor and eventually founded Lehigh's CS department.  He joined UCCS in 2003 as an El Pomar Professor.  Over his career, he has won multiple teaching awards, research/innovation awards, best paper awards, best reviewer awards, and IEEE service awards; He is a member of the IEEE Golden Core and has been an IEEE Distinguished Lecturer, and in 2017 was elected as an IEEE Fellow. He was a co-founder of the Computer Vision Foundation and very active in organizing/managing Computer Vision conferences.  On the education side, Dr.\@ Boult is the founder, primary architect, and co-director of the world's first and only Bachelor of Innovation™ Family of Degrees at UCCS.  This awarding family of degrees combines a core of innovation and entrepreneurship with a significant multi-year "team emphasis" and all the rigor of bachelor degrees in their fields, serving 100s or students across 22 different majors spanning four colleges.
\end{IEEEbiography}

\newpage

\begin{IEEEbiography}[{\includegraphics[width=1in,height=1.25in,clip,keepaspectratio]{\changedir{bio_photos/}Jacques_Wainer}}]{Jacques Wainer}
is a professor at the Computing Institute at the State University of Campinas. He received his PH.D in Computer Science in 1991 from the Pennsylvania State University. He has worked in areas such as collaborative computing, medical informatics, bibliometrics, social impacts of computers, and others. Currently, his main focus of research is machine learning.
\end{IEEEbiography}

\begin{IEEEbiography}[{\includegraphics[width=1in,height=1.25in,clip,keepaspectratio]{\changedir{bio_photos/}Rocha}}]{Anderson Rocha}
  is an Associate Professor at the Institute of Computing, University of Campinas (Unicamp), Brazil, since 2009.
  He obtained his bachelor's degree (2003) in Computer Science at Federal University of Lavras, Brazil.
  He obtained his M.Sc. (2006) and Ph.D. (2009) in Computer Science at Unicamp.
  His research interests include Machine Learning, Reasoning for Complex Data, and Digital Forensics.
  He is an IEEE Senior member and the Chair of the IEEE Information Forensics and Security Technical Committee (IFS-TC) for the 2019-2020 term. He has been actively involved in the organization of several events such as ICIP, ICASSP, WIFS in the past years and has been an Associate Editor of several journals such as the IEEE Security \& Privacy, IEEE Signal Processing Letters, and the IEEE Transactions on Information Forensics and Security, and a Senior Editor for the IEEE Signal Processing Letters. He is a Microsoft, Google, and Tan Chi Tuan Faculty Fellow, honorable recognitions for his research contributions.
\end{IEEEbiography}




\end{document}